\documentclass{article} 
\usepackage{iclr2025_conference,times}

\usepackage{amsmath,amsfonts,bm}

\def\eqref#1{equation~\ref{#1}}

\def\1{\bm{1}}

\DeclareMathAlphabet{\mathsfit}{\encodingdefault}{\sfdefault}{m}{sl}
\SetMathAlphabet{\mathsfit}{bold}{\encodingdefault}{\sfdefault}{bx}{n}

\usepackage{hyperref}
\usepackage{url}
\usepackage{algorithm}
\usepackage{algpseudocode}
\usepackage{multirow} 
\usepackage{array}
\usepackage{graphicx} 
\usepackage{booktabs}

\iclrfinalcopy 
\begin{document}

{\LARGE \scshape \title{Looped World Models} \par}
\begin{center}

\maketitle
\vspace{-2em}

{\large \textbf{FaceMind Research Asia}\par\includegraphics[height=1.2cm]{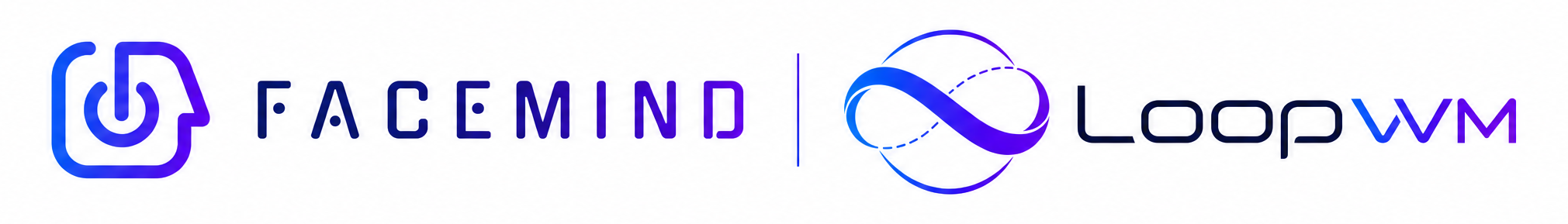}}

{\normalsize
\textbf{Leading Contributors}\par Hongyuan Adam Lu\textsuperscript{*}\quad Z.L.\quad Victor Wei\par
\vspace{0.4em}
\textbf{Core Contributors}\par
Qun Zhang \quad Jinrui Zeng \quad Bowen Cao \quad Lingwei Meng \quad Mocheng Li \par
Zezhong Wang \quad Haonan Yin \quad Naifu Xue \quad Minyu Chen \quad Cenyuan Zhang \par
Zefan Zhang \quad Hao Wei \quad Jiawei Zhou \quad Haoran Xu \quad Hao Yang \par
Ronglai Zuo \quad Tongda Xu \quad Yonghao Li \quad Jian Chen \quad Hebin Wang \par
Zeyu Gao \quad Yang Li \quad Wei Zhao \quad Qimin Zhong \quad Siqi Liu \par
Yumeng Zhang \quad Leyan Cui \quad Zhangyu Wang \quad Wai Lam
}

\end{center}

\begin{abstract}
Current world models face a fundamental tension: faithful long-horizon simulation demands deep computation, but deeper models are expensive to deploy and prone to compounding errors. We resolve this by introducing Looped World Models (LoopWM), which are the first looped architectures for world modelling. Our method iteratively refines latent environment states through a parameter-shared transformer block. This yield up to 100× parameter efficiency over conventional approaches with adaptive computation that automatically scales depth to match the complexity of each prediction step. Orthogonal to scaling model size and training data, LoopWM establishes iterative latent depth as a new scaling axis for world simulation, which might significantly push the community forward.

\end{abstract}

\begin{figure*}[h]
\centering
\includegraphics[width=11.5cm]{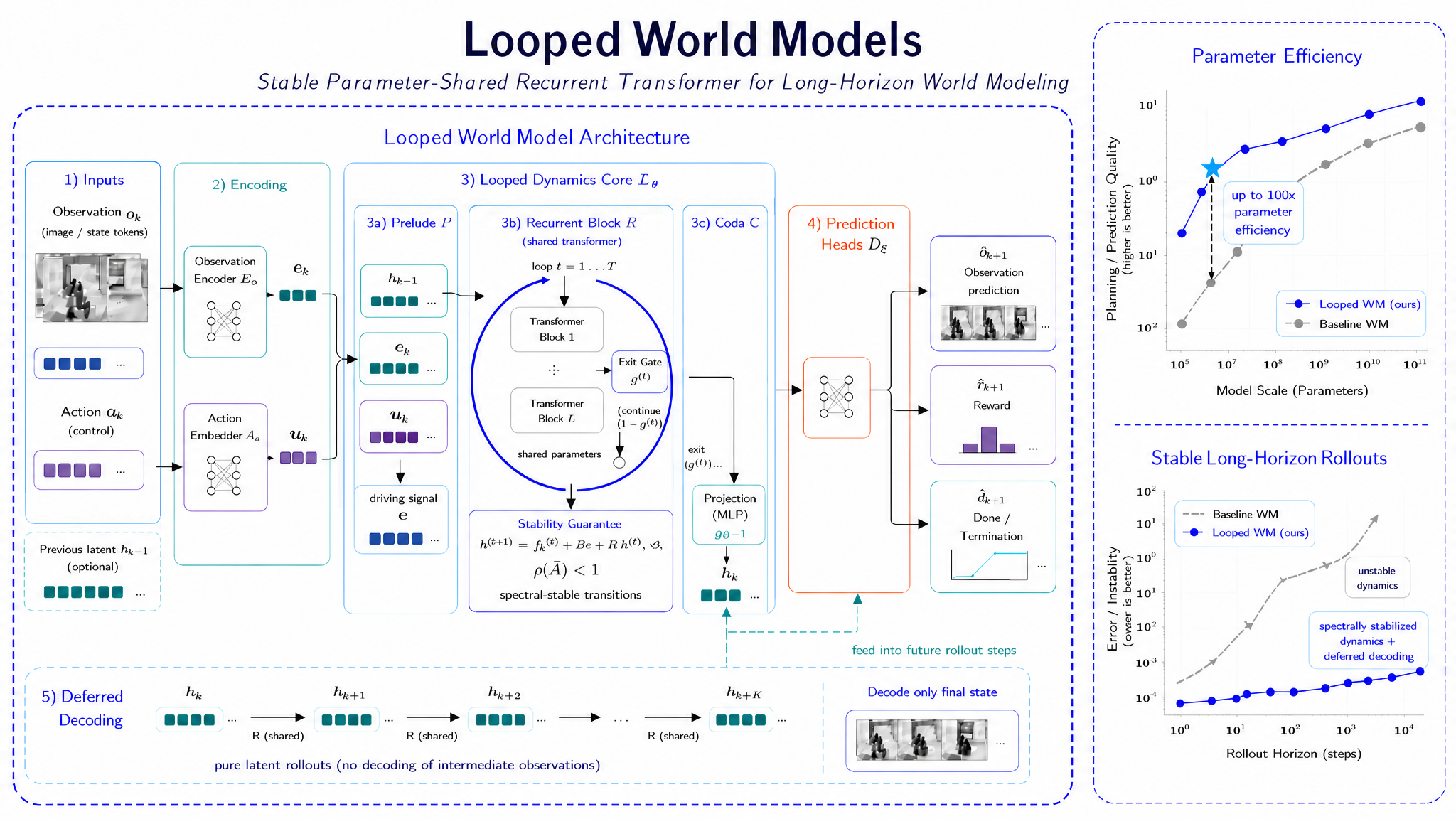}
\caption{The overall framework of our proposed Looped World Models (LoopWM).}
\label{fig: method}
\end{figure*}

\section{Introduction}
World models (WM) learn to predict how an environment evolves in accordance with actions. WM has become a cornerstone of sample-efficient reinforcement learning and embodied intelligence \citep{ha2018worldmodels, hafner2019planet, kaiser2020simple}. Remarkably, the Deep Planning Network (PlaNet) is a WM \citep{hafner2019planet} first demonstrated that agents can learn latent dynamics entirely from pixels and plan via online optimisation. This establishes the recurrent state-space model (RSSM) as a foundational architecture for world modelling. The Dreamer family of models then \citep{hafner2020dreamerv1, hafner2021dreamerv2, hafner2025dreamerv3} progressively refined this approach, culminating in DreamerV3 \citep{hafner2025dreamerv3}. DreamerV3 masters over 150 different tasks with a single set of hyperparameters. Seeking to leverage the representational power of transformers, subsequent work replaced or augmented the recurrent backbone. IRIS \citep{micheli2023iris} showed that an autoregressive transformer over discrete latent tokens can serve as a highly data-efficient world model. TransDreamer \citep{chen2022transdreamer} introduced a Transformer State-Space Model for tasks demanding long-range memory. $\Delta$-IRIS \citep{micheli2024deltairis} improved efficiency via context-aware delta tokenisation. DIAMOND \citep{alonso2024diamond} demonstrated that diffusion models can produce visually faithful world simulations, and EMERALD \citep{burchi2025emerald} achieved state-of-the-art Crafter performance by combining masked generative transformers with spatial latent states. At a larger scale, Sora \citep{openai2024sora} and Genie \citep{bruce2024genie, deepmind2025genie3} demonstrated that video generation models and generative interactive environments can serve as general-purpose world simulators. And multiple surveys have charted the rapid expansion of world models into autonomous driving \citep{wang2025wmadsurvey}, embodied AI \citep{zhu2025embodiedwmsurvey}, and video generation \citep{cho2024sorasurvey, wang2025mechanisticvideo}. 

Despite this progress, faithful long-horizon simulation often requires deep or iterative computation. This is because physical dynamics unfold through repeated application of governing laws, whereas conventional fixed-depth architectures allocate the same amount of computation to every transition regardless of its difficulty. There are two typical failure modes. First, prediction errors cause trajectory quality to degrade rapidly over extended horizons across rollout steps \citep{xiao2019compoundingerror, talvitie2017self, lambert2022survey}. Second, scaling model depth to combat this degradation proportionally usually increases parameter count and inference cost, which then makes real-time deployment on resource-constrained platforms prohibitively expensive \citep{wang2025wmadsurvey, hafner2025dreamerv3}.

A parallel line of research has explored \emph{looped transformer architectures} (LM). In LM, a shared set of transformer blocks is applied recurrently to the same latent representation. Such a concept was first proposed as the Universal Transformer \citep{dehghani2019universal}, which introduced weight-sharing across depth with an adaptive halting mechanism inspired by Adaptive Computation Time \citep{graves2016act}. Early theoretical work shows that LM can simulate arbitrary iterative algorithms. This includes gradient descent, Newton's method, and dynamic programming, with constant parameter count \citep{giannou2023programmable}, and they achieve comparable in-context learning performance to standard transformers while using less than 10\% of the parameters \citep{yang2023looped}. ALBERT \citep{lan2020albert} shows the practical viability of cross-layer parameter sharing for language representation learning. MoEUT \citep{csordas2024moeut} combined mixture-of-experts with universal transformers.

More recently, looped architectures have been scaled to practical language models with promising results. \citet{zhu2025ouro} demonstrated that a looped language model can achieve about 2 to 3$\times$ parameter efficiency through iterative latent computation. \citet{geiping2025rdm} showed that recurrent-depth models can scale test-time compute by simply increasing the number of loop iterations at inference. \citet{fan2024lengthgen} demonstrated that looped transformers with adaptive stopping significantly improve length generalisation.  \citet{saunshi2025latentthoughts} provided theoretical and empirical evidence that looped models implicitly generate latent thoughts. \citet{jeddi2025loopformer} introduced elastic-depth training with shortcut modulation for budget-conditioned latent reasoning. \citet{bae2025mor} proposed per-token dynamic recursive depth allocation within a single recursive transformer. \citet{prairie2026parcae} addressed the training instability of looped models by recasting the looped forward pass as a nonlinear dynamical system over the residual stream and constraining the spectral norm of the state-transition matrix through a negative-diagonal parameterisation. Most recently, Hyperloop Transformers \citep{zeitoun2026hyperloop} augmented the looped block with matrix-valued hyper-connected residual streams. This outperforms depth-matched standard transformers at half the parameter count. These developments connect looped transformers to a broader family of depth-continuous and implicit-layer models, including Neural ODEs \citep{chen2018neuralode} and Deep Equilibrium Models \citep{bai2019deq}, which likewise iterate a shared function toward a fixed point.

However, \emph{all} of the above looped-architecture works have been developed and evaluated exclusively in the context of language modelling. Looped World Models (LoopWM) remain entirely unexplored.

We propose that looped transformers are a promising backbone for world models because they introduce an explicit iterative refinement mechanism while reusing parameters across depth. At a high level, environment dynamics can often be viewed as repeated application of a shared transition law, which motivates modelling a single-step transition through repeated application of a shared latent update operator. This correspondence is conceptual rather than exact, as the inner loop is not meant to represent physical time directly but to perform iterative refinement of a latent transition estimate. To improve the numerical stability of this recurrent computation, we adopt a spectrally constrained state-retention parameterisation inspired by looped architectures. This construction ensures that the linear retention component remains contractive, which helps keep recurrent latent updates bounded as the number of inner-loop iterations increases. Structurally, environment dynamics are themselves an iterative process: a state $s_t$ evolves to $s_{t+1}$ through the repeated application of (approximately) stationary physical laws. The looped transformer's computation graph, where a shared function $f_\theta$ is applied recurrently to a latent state $h$,
\begin{equation}
  h_{t+1} = \bar{A}\, h_t + \bar{B}\, e + \bar{\mathcal{R}}(h_t, e),
  \label{eq:looped_update}
\end{equation} 
with $\bar{A}$ governing state retention, $\bar{B}$ controlling input injection, and $\bar{\mathcal{R}}$ subsuming the transformer nonlinearities \citep{prairie2026parcae} is directly isomorphic to this dynamics structure. Stability is guaranteed by parameterizing the continuous-time matrix as $A := \mathrm{diag}(-\exp(\mathbf{a}))$ with learnable $\mathbf{a}$, and discretizing via zero-order hold,
\begin{equation}
  \bar{A} = \exp(\Delta\, A),
  \label{eq:discretization}
\end{equation}
which constrains all eigenvalues of $\bar{A}$ to the interval $(0, 1)$, ensuring bounded residual dynamics regardless of rollout length
\citep{prairie2026parcae}.

Practically, the parameter efficiency of looped architectures is uniquely valuable for world models, because long-horizon rollouts require executing the dynamics model hundreds or thousands of times in sequence; a model that achieves the predictive quality of a much larger network with a fraction of the parameters yields compounding savings across every rollout step. Furthermore, the adaptive-depth property of looped models, allocating more iterations to complex transitions (e.g., collisions, contact events) and fewer to simple dynamics (e.g., free flight), maps directly onto the non-uniform computational demands of physical simulation. In the most favourable cases, where simple state transitions require only a single loop iteration compared to the full forward pass of a conventional fixed-depth model, this adaptive mechanism can substantially reduce average inference cost relative to a fixed-depth baseline. The magnitude of this reduction depends on the distribution of transition difficulty, the minimum useful loop depth, and the overhead of the exit mechanism.

In this work, we introduce \textbf{Looped World Models} (LoopWM), the first looped transformer architectures for environment simulation and dynamics prediction. Our approach combines a parameter-shared recurrent transformer block with spectrally-constrained residual dynamics, enabling provably stable state transitions across arbitrary rollout lengths. We demonstrate that Looped World Models achieve competitive or superior predictive accuracy to existing world model architectures while using significantly fewer parameters, maintain stable rollouts over substantially longer horizons, and support test-time adaptive computation that automatically matches computational depth to transition complexity. We also integrate residual connections to improve model performance. Our results establish iterative latent depth as a previously unexplored and highly effective scaling axis for world models, orthogonal to both model size and training data.

\section{Related Work}
\subsection{World Models for Reinforcement Learning and Embodied AI}

The idea of learning an internal model of environment dynamics dates back to early work on mental simulation and forward models in cognitive science and control theory. In deep reinforcement learning, \citet{ha2018worldmodels} proposed learning a compressed spatial and temporal representation of the environment using a variational autoencoder and an RNN, training a compact policy entirely within the learned ``dream.'' PlaNet \citep{hafner2019planet} formalised this via a latent dynamics model (RSSM) that plans directly in latent space from pixel observations. SimPLe \citep{kaiser2020simple} demonstrated model-based Atari play by training a video-prediction model as a learned simulator. MuZero \citep{schrittwieser2020muzero} showed that a learned dynamics model with Monte-Carlo tree search can master board games and Atari without access to the ground-truth rules.

The Dreamer family \citep{hafner2020dreamerv1, hafner2021dreamerv2, hafner2025dreamerv3} progressively refined the RSSM-based world model, culminating in DreamerV3 \citep{hafner2025dreamerv3}. They achieve human-level performance across over 150 diverse tasks with a single set of hyperparameters. Transformer-based world models subsequently emerged: IRIS \citep{micheli2023iris} replaced the recurrent backbone with an autoregressive transformer over discrete tokens; TransDreamer \citep{chen2022transdreamer} introduced a Transformer State-Space Model for memory-demanding tasks; $\Delta$-IRIS \citep{micheli2024deltairis} improved tokenization efficiency via context-aware delta encoding; DIAMOND \citep{alonso2024diamond} leveraged diffusion models to produce visually faithful world simulations; and EMERALD \citep{burchi2025emerald} achieved state-of-the-art Crafter performance using masked generative transformers over spatial latent states.

At a larger scale, video generation models have been cast as world simulators. OpenAI's Sora \citep{openai2024sora} demonstrated long-form video generation with emergent 3D consistency, while Genie \citep{bruce2024genie} and Genie~3 \citep{deepmind2025genie3} showed that text-conditioned generative models can produce interactive, explorable environments. Several surveys chart the rapid expansion of world models into autonomous driving \citep{wang2025wmadsurvey, wang2024wminitialsurvey}, embodied AI \citep{zhu2025embodiedwmsurvey}, and video generation \citep{cho2024sorasurvey, wang2025mechanisticvideo}.

A persistent challenge across all these approaches is \emph{compounding prediction error}: small inaccuracies at each rollout step accumulate exponentially over long horizons, degrading trajectory fidelity \citep{xiao2019compoundingerror, talvitie2017self, lambert2022survey}. Various mitigation strategies have been proposed, including short-horizon re-planning, self-correcting models \citep{talvitie2017self}, and physics-informed architectures \citep{pinwm2025, prophy2025}, yet the fundamental tension between computational depth and rollout stability remains unresolved by existing architectures.

\subsection{Looped and Recurrent-Depth Transformer Architectures}

Looped transformers reuse a shared set of transformer blocks across depth, decoupling effective computation from parameter count. The Universal Transformer \citep{dehghani2019universal} first proposed this idea, combining weight sharing with Adaptive Computation Time (ACT) \citep{graves2016act} for input-dependent halting. ALBERT \citep{lan2020albert} demonstrated the practical viability of full cross-layer parameter sharing in BERT-scale models.

Theoretical analyses subsequently established the computational power of looped transformers. \citet{giannou2023programmable} proved that looped transformers can simulate arbitrary programs, functioning as programmable computers with constant parameter count.  \citet{yang2023looped} showed that looped transformers match standard transformer performance on in-context learning while using less than 10\% of the parameters. \citet{fan2024lengthgen} demonstrated significant length generalisation improvements through adaptive loop counts. \citet{saunshi2025latentthoughts} provided both theoretical and empirical evidence that looped models implicitly generate ``latent thoughts,'' enabling reasoning beyond their apparent depth. At a practical scale, Ouro \citep{zhu2025ouro} trained looped language models (LoopLMs) through the full modern LLM pipeline with pre-training, instruction tuning, and RLHF, achieving 2--3$\times$ parameter efficiency with entropy-regularised adaptive computation. \citet{geiping2025rdm} demonstrated that recurrent-depth models (RDMs) scale test-time compute by increasing loop count at inference, following predictable quality improvements. \citet{pappone2025twoscale} analysed the geometry of latent dynamics in recurrent-depth transformers, identifying two-scale structure with fast intra-loop and slow inter-token dynamics. LoopFormer \citep{jeddi2025loopformer} introduced elastic-depth training with shortcut modulation for budget-conditioned reasoning. Mixture-of-Recursions \citep{bae2025mor} proposed per-token dynamic depth allocation within a single recursive framework. MoEUT \citep{csordas2024moeut} combined mixture-of-experts with universal transformers to balance specialisation and sharing.

\subsection{Adaptive Computation and Early Exit}

Allocating variable computation to inputs of differing complexity has been studied across multiple paradigms. \citet{graves2016act} introduced Adaptive Computation Time for RNNs, allowing per-step halting decisions. The early exit literature \citep{teerapittayanon2016branchynet, bolukbasi2017adaptive, earlyexit2025survey} enables inference to terminate at intermediate layers when confidence is sufficient. In the context of looped transformers, adaptive depth takes a particularly natural form: the model can halt after any number of loop iterations. Ouro \citep{zhu2025ouro} introduced entropy-regularised early exit, where a token exits the loop when its prediction entropy drops below a learned threshold. \citet{geiping2025rdm} trained with stochastic depth sampling (Poisson-distributed loop counts) to induce robustness to variable test-time depth. LoopFormer \citep{jeddi2025loopformer} conditioned on a continuous ``time budget'' during training, enabling fine-grained compute allocation at inference. \citet{pappone2025twoscale} proposed acceleration-based exit rules using second-order differences of hidden states. For world models specifically, adaptive computation is highly attractive: simple state transitions (e.g., free flight, static scenes) demand minimal processing, while complex events (e.g., multi-body collisions, contact dynamics) require deeper iterative refinement. To the best of our knowledge, no prior work has proposed adaptive-depth looped architectures with world modelling.

\section{Looped World Model}
We present Looped World Models, a latent dynamics architecture that combines the iterative computation of looped transformers with the action-conditioned state prediction required for world modelling. Our design follows three principles: (i)~structural alignment between the model's computation graph and the iterative nature of physical dynamics, (ii)~provable stability of latent state transitions across arbitrary rollout lengths, and (iii)~adaptive computational depth that matches the complexity of each transition. We describe the overall architecture (\S\ref{sec:architecture}), the stabilised looped dynamics core (\S\ref{sec:dynamics}), the training objective (\S\ref{sec:training}), and the adaptive early-exit mechanism for inference (\S\ref{sec:earlyexit}).

\subsection{Overall Architecture}
\label{sec:architecture}

At each environment time step $k$, the agent receives an observation $o_k \in \mathcal{O}$ and selects an action $a_k \in \mathcal{A}$. The goal of the world model is to predict the next latent state, from which future observations, rewards, and termination signals can be decoded. Our architecture consists of four modules:

\paragraph{Observation Encoder $\mathcal{E}_\phi$.} A convolutional (or vision-transformer-based) encoder maps the raw observation $o_k$ into a latent embedding $e_k = \mathcal{E}_\phi(o_k) \in \mathbb{R}^{d}$.

\paragraph{Action Embedder $\mathcal{A}_\psi$.} The action $a_k$ is projected into the same latent space via a learned embedding $u_k = \mathcal{A}_\psi(a_k) \in \mathbb{R}^{d}$.

\paragraph{Looped Dynamics Core $\mathcal{L}_\theta$.} This is the central contribution of our architecture. The dynamics core takes the previous latent state $h_{k-1}$, the current observation embedding $e_k$, and the action embedding $u_k$, and produces the next latent state $h_k$ through $T$ iterations of a parameter-shared transformer block with spectrally-constrained residual dynamics. We describe this module in detail in \S\ref{sec:dynamics}.

\paragraph{Prediction Heads $\mathcal{D}_\xi$.} A set of lightweight MLPs decode the latent state $h_k$ into: (i)~a reconstructed observation $\hat{o}_{k+1}$ or its latent target, (ii)~a predicted reward $\hat{r}_k$, and (iii)~a predicted continuation flag $\hat{c}_k$. These heads follow the standard design of prior latent world models \citep{hafner2020dreamerv1, hafner2021dreamerv2, hafner2025dreamerv3}.

The full forward pass at environment step $k$ proceeds as:
\begin{equation}
  e_k = \mathcal{E}_\phi(o_k), \quad
  u_k = \mathcal{A}_\psi(a_k), \quad
  h_k = \mathcal{L}_\theta(h_{k-1},\, e_k,\, u_k), \quad
  (\hat{o}_{k+1},\, \hat{r}_k,\, \hat{c}_k)
    = \mathcal{D}_\xi(h_k).
  \label{eq:full_forward}
\end{equation}

During imagination-based training of the policy (as in Dreamer \citep{hafner2020dreamerv1}), the encoder is bypassed: the dynamics core autoregressively rolls out latent trajectories using only actions sampled from the policy network, i.e., $h_{k+1} = \mathcal{L}_\theta(h_k,\, \mathbf{0},\, u_k)$, where observation injection is omitted or replaced by the model's own prediction.

\subsection{Looped Dynamics Core with Spectral Stability}
\label{sec:dynamics}

The dynamics core is the heart of our architecture. Following the prelude recurrent coda design \citep{geiping2025rdm, prairie2026parcae, zeitoun2026hyperloop}, we partition the dynamics core into three blocks:

\paragraph{Prelude $\mathcal{P}$.} A small stack of $L_\mathcal{P}$ transformer layers processes the concatenation of the previous latent state, the observation embedding, and the action embedding to produce the conditioning signal:
\begin{equation}
  e = \mathrm{LN}\!\left(
    \mathcal{P}\!\left(
      [h_{k-1};\, e_k;\, u_k]
    \right)
  \right) \in \mathbb{R}^{d},
  \label{eq:prelude}
\end{equation}
where $\mathrm{LN}(\cdot)$ denotes layer normalization. The normalisation of $e$ follows the Parcae design \citep{prairie2026parcae} and prevents input magnitude from inducing late-stage loss spikes.

\paragraph{Recurrent Block $\mathcal{R}$.}
A stack of $L_\mathcal{R}$ transformer layers with shared parameters is applied iteratively for $T$ loops. The hidden state is initialised as $h^{(0)} \sim \mathcal{N}(0, \sigma^2 I)$ (or, for temporal rollouts, from the previous time step's final hidden state). At each loop iteration $t = 0, 1, \ldots, T{-}1$, the update rule is:
\begin{equation}
  h^{(t+1)} = \bar{A}\, h^{(t)}
             + \bar{B}\, e
             + \bar{\mathcal{R}}(h^{(t)},\, e),
  \label{eq:recurrent_update}
\end{equation}
where $\bar{A} \in \mathbb{R}^{d \times d}$ is the state-retention matrix controlling how much of the previous hidden state is preserved, $\bar{B} \in \mathbb{R}^{d \times d}$ is the input-injection matrix controlling the influence of the conditioning signal $e$, and $\bar{\mathcal{R}}$ subsumes the nonlinear transformer operations (multi-head attention and feed-forward layers) applied to the residual stream. The key distinction from conventional fixed-depth transformers is that the parameters of $\mathcal{R}$ are \emph{shared across all $T$ iterations}, making the computational depth independent of the parameter count.

\paragraph{Spectral Stability Constraint.} To guarantee that the latent state does not explode regardless of the number of loop iterations $T$ (which is critical for long-horizon rollouts in world modelling), we constrain the spectral norm of $\bar{A}$ to be strictly less than 1. Following Parcae \citep{prairie2026parcae}, we parameterize $\bar{A}$ through discretization of a continuous-time negative diagonal matrix:
\begin{align}
  A &:= \mathrm{diag}\!\left(-\exp(\mathbf{a})\right),
      \quad \mathbf{a} \in \mathbb{R}^{d}~\text{(learnable)},
      \label{eq:A_continuous} \\
  \bar{A} &= \exp(\Delta \cdot A),
      \quad \Delta \in \mathbb{R}^{d}_{>0}~\text{(learnable)}.
      \label{eq:A_discrete}
\end{align}
Since $A$ has strictly negative diagonal entries, $\Delta \cdot A$ has strictly negative entries, and $\exp(\cdot)$ maps these to the interval $(0, 1)$. Consequently, $\bar{A}$ is a diagonal matrix with all entries in $(0, 1)$, guaranteeing $\rho(\bar{A}) < 1$. This constraint holds by construction throughout training; no gradient clipping, post-hoc normalisation, or sensitive hyperparameter tuning is required.

The input-injection matrix is similarly discretised as $\bar{B} = \Delta \cdot B$ with unconstrained $B$, but we apply layer normalisation to $e$ (Eq.~\ref{eq:prelude}) to bound the injected signal's magnitude.

\paragraph{Coda $\mathcal{C}$.} A final stack of $L_\mathcal{C}$ transformer layers (with separate, non-shared parameters) processes the terminal hidden state $h^{(T)}$ through a learned projection:
\begin{equation}
  h_k = \mathcal{C}(C\, h^{(T)}),
  \label{eq:coda}
\end{equation}
where $C \in \mathbb{R}^{d_c \times d}$ optionally adapts the embedding dimension. The output $h_k$ is then passed to the prediction heads and carried forward as the initial state for the next environment time step.

\paragraph{Cross-Timestep State Propagation.} A distinctive property of our architecture is that the terminal hidden state $h^{(T)}$ from environment step $k$ can serve as the initialization $h^{(0)}$ for step $k{+}1$, enabling a dual-loop structure: the \emph{inner loop} (iterations of $\mathcal{R}$) refines the latent state within a single transition, while the \emph{outer loop} (sequential environment steps) propagates information across time. The spectral constraint on $\bar{A}$ ensures that both loops remain bounded, encouraging continuity while keeping propagated hidden states numerically well behaved.

\subsection{Training Objective}
\label{sec:training}

\paragraph{Variable-Depth Training.} We train with stochastic loop depth. At each training step, the loop count $T$ is sampled from a Poisson distribution with learnable mean $\mu_{\mathrm{rec}}$:
\begin{equation}
  T \sim \mathrm{Poisson}(\mu_{\mathrm{rec}}).
  \label{eq:depth_sampling}
\end{equation}
We sample $T$ independently \emph{per sequence} within each micro-batch, rather than per micro-batch as in prior work \citep{geiping2025rdm}. This reduces variance in the training objective and empirically eliminates most loss spikes.

\paragraph{World Model Loss.}
The overall training objective combines observation prediction, reward prediction, and continuation prediction:
\begin{equation}
  \mathcal{L}_{\mathrm{wm}} =
    \mathbb{E}_{T \sim \mathrm{Poisson}(\mu_{\mathrm{rec}})}
    \left[
      \sum_{k=1}^{K}
      \left(
        \underbrace{\mathcal{L}_{\mathrm{obs}}(o_{k+1},\, \hat{o}_{k+1})}
          _{\text{observation loss}}
        + \lambda_r\,
          \underbrace{\mathcal{L}_{\mathrm{rew}}(r_k,\, \hat{r}_k)}
          _{\text{reward loss}}
        + \lambda_c\,
          \underbrace{\mathcal{L}_{\mathrm{cont}}(c_k,\, \hat{c}_k)}
          _{\text{continuation loss}}
      \right)
    \right],
  \label{eq:wm_loss}
\end{equation}
where $K$ is the sequence length, $\lambda_r$ and $\lambda_c$ are balancing coefficients, and the specific form of $\mathcal{L}_{\mathrm{obs}}$ depends on the observation space (e.g., MSE for continuous states, cross-entropy for discrete tokens). Backpropagation through the loop iterations is truncated at $\mu_{\mathrm{bwd}} = \lceil \mu_{\mathrm{rec}} / 2 \rceil$ steps to limit memory cost.

\paragraph{Entropy-Regularised Adaptive Depth.} When adaptive early exit is enabled (see \S\ref{sec:earlyexit}), we augment the loss with an entropy-regularisation term that prevents the exit gate from collapsing to trivial solutions (always exiting at the first iteration or never exiting). The regularisation takes the form:
\begin{equation}
  \mathcal{L}_{\mathrm{ent}} =
    -\alpha\, \mathbb{E}
    \left[
      \sum_{t=1}^{T} H\!\left(g^{(t)}\right)
    \right],
  \label{eq:entropy_reg}
\end{equation}
where $g^{(t)} \in [0, 1]$ is the exit probability at loop iteration $t$, $H(\cdot)$ denotes binary entropy, and $\alpha$ is a regularization coefficient. The total training loss is $\mathcal{L} = \mathcal{L}_{\mathrm{wm}} + \mathcal{L}_{\mathrm{ent}}$.

\subsection{Adaptive Early Exit for Inference}
\label{sec:earlyexit}

At inference time, the looped dynamics core can adaptively terminate the inner loop early for transitions that converge quickly, and allocate additional iterations to complex transitions. We implement this via a lightweight exit gate, a single-layer MLP followed by a sigmoid:
\begin{equation}
  g^{(t)} = \sigma\!\left(
    \mathbf{w}_g^\top\, h^{(t)} + b_g
  \right),
  \label{eq:exit_gate}
\end{equation}
where $\mathbf{w}_g \in \mathbb{R}^{d}$ and $b_g \in \mathbb{R}$ are learned parameters. At each loop iteration $t$, if $g^{(t)}$ exceeds a threshold $\tau$, the loop terminates and $h^{(t)}$ is used as the final hidden state. This mechanism is complementary to the convergence-based exit criteria studied by  \citet{pappone2025twoscale}, which halt when the second-order difference $\|h^{(t)} - 2h^{(t-1)} + h^{(t-2)}\|$ falls below a threshold.

In the world-modelling setting, adaptive exit yields particularly large savings. Consider a 100-layer fixed-depth baseline: for a simple free-flight trajectory segment, our model may exit after a single loop of $L_\mathcal{R}$ layers (e.g., 4 layers), reducing inference FLOPs by a factor of $\sim$25$\times$ for that step. Over a long rollout containing many simple transitions interspersed with occasional complex events, the aggregate FLOPs reduction can reach up to two orders of magnitude compared to a fixed-depth model of equivalent quality.

The maximum loop count $T_{\max}$ at inference time can also exceed the training-time mean $\mu_{\mathrm{rec}}$, enabling test-time compute scaling: the model produces progressively refined predictions as more iterations are allocated.

\subsection{Deferred Decoding: Action-Conditioned Latent Rollout}
\label{sec:deferred_decoding}

\subsubsection{Motivation}

In standard world models~\citep{hafner2020dreamerv1,hafner2021dreamerv2,hafner2025dreamerv3}, the prediction heads $\mathcal{D}_\xi$ are applied at every environment step $k$ to produce intermediate observation reconstructions $\hat{o}_{k+1}$, reward predictions $\hat{r}_k$, and continuation signals $\hat{c}_k$. This per-step decoding introduces two inefficiencies: (i) it forces the latent state to allocate representational capacity to pixel-level reconstruction at every intermediate step, even when only the final prediction matters for planning; (ii) it prevents the dynamics core from performing uninterrupted latent reasoning across a multi-step action sequence.

Recent work in language modelling has demonstrated that deferring decoding to the end of a latent reasoning process---allowing the model to \emph{encode, think, then decode}---substantially improves reasoning quality~\citep{koishekenov2025etd,geiping2025rdm,saunshi2025latentthoughts}. MuZero~\citep{schrittwieser2020muzero} similarly operates entirely in latent space without observation reconstruction, predicting only value, reward, and policy.  Dreamer's own ``imagination'' rollouts~\citep{hafner2020dreamerv1} propagate latent states without re-encoding real observations, yet still apply reward and value heads at each step.

We propose \textbf{Deferred Decoding}, a modification to the Looped World Model that eliminates all intermediate observation decoding during multi-step rollouts. Given a sequence of ground-truth or planned actions, the model injects each action into the looped dynamics core and advances the latent state purely in the continuous hidden space.  Observation, reward, and continuation predictions are produced \emph{only at the final step}, reducing computation and encouraging the latent trajectory to encode temporally extended, action-relevant structure rather than per-step visual detail.

\subsubsection{Formulation}

Consider a planning or evaluation horizon of $K$ steps. Let $h_0$ be the initial latent state (obtained from encoding a real observation $o_0$ through the encoder $\mathcal{E}_\phi$ and the prelude block of the looped dynamics core), and let $(a_0, a_1, \ldots, a_{K-1})$ be a sequence of actions.

\paragraph{Standard per-step decoding (baseline).}
At each step $k = 0, 1, \ldots, K-1$, the baseline model performs:
\begin{align}
  u_k &= \mathcal{A}_\psi(a_k), \label{eq:dd_action} \\
  h_{k+1} &= \mathcal{L}_\theta(h_k, u_k),  \label{eq:dd_dynamics} \\
  (\hat{o}_{k+1},\, \hat{r}_k,\, \hat{c}_k) &= \mathcal{D}_\xi(h_{k+1}),
  \label{eq:dd_decode_all}
\end{align}
where $\mathcal{L}_\theta$ denotes the full looped dynamics core (prelude, $T$-step recurrent block, coda) described in Section~\ref{sec:dynamics}. This yields $K$ sets of decoded predictions.

\paragraph{Deferred decoding.}
We replace the per-step decoding with a \emph{decode-free latent rollout} followed by a single terminal decoding:
\begin{align}
  u_k &= \mathcal{A}_\psi(a_k),
    &k &= 0, 1, \ldots, K-1, \label{eq:dd2_action} \\
  h_{k+1} &= \mathcal{L}_\theta^{\mathrm{core}}(h_k, u_k),
    &k &= 0, 1, \ldots, K-1, \label{eq:dd2_dynamics} \\
  (\hat{o}_{K},\, \hat{r}_{K},\, \hat{c}_{K}) &=
    \mathcal{D}_\xi(h_K). \label{eq:dd2_final_decode}
\end{align}
The key difference is that Eqs.~\eqref{eq:dd2_action}, \eqref{eq:dd2_dynamics} are applied $K$ times \emph{without invoking $\mathcal{D}_\xi$}, and the decoder is called exactly once at step $K$. Between steps, the model ingests a new action embedding $u_k$ and advances the latent state through the looped recurrent block, but produces no intermediate observation, reward, or continuation output.

\paragraph{Interaction between inner and outer loops.}
Recall from Section~\ref{sec:dynamics} that each invocation of $\mathcal{L}_\theta^{\mathrm{core}}$ itself involves $T$ inner-loop iterations of the shared transformer block.  With deferred decoding, the overall computation becomes a \emph{nested loop}:
\begin{itemize}
  \item \textbf{Outer loop} (action steps): $k = 0, \ldots, K-1$.
    At each step, the action $u_k$ is injected.
  \item \textbf{Inner loop} (latent refinement): $t = 0, \ldots, T-1$.
    Within each action step, the recurrent block refines $h$ via
    $h^{(t+1)} = \bar{A}\,h^{(t)} + \bar{B}\,[u_k;\, h_k]
    + \bar{\mathcal{R}}(h^{(t)}, u_k)$
    with spectral-norm-constrained $\bar{A}$.
\end{itemize}
The total effective depth is $K \times T$ shared-parameter transformer applications, but only one forward pass through the decoder.

\subsubsection{Training Objective for Deferred Decoding}
\label{sec:dd_training}

Training the deferred-decoding variant requires the model to maintain accurate latent representations across $K$ action-conditioned transitions \emph{without} intermediate reconstruction supervision. We define a \textbf{terminal prediction loss} and a \textbf{latent trajectory regularizer}.

\paragraph{Terminal prediction loss.}
Given a training trajectory $(o_0, a_0, o_1, a_1, \ldots, o_K)$ where all intermediate actions are ground-truth, the model encodes $o_0$ to obtain $h_0$, performs $K$ latent transitions via Eqs.~\eqref{eq:dd2_action}, \eqref{eq:dd2_dynamics}, then decodes $h_K$:
\begin{equation}
  \mathcal{L}_{\mathrm{terminal}} =
    \lambda_o\,\ell_{\mathrm{obs}}(\hat{o}_K, o_K)
    + \lambda_r\,\ell_{\mathrm{rew}}(\hat{r}_K, r_{K-1})
    + \lambda_c\,\ell_{\mathrm{cont}}(\hat{c}_K, c_{K-1}),
  \label{eq:dd_loss_terminal}
\end{equation}
where $\ell_{\mathrm{obs}}$ may be a reconstruction loss (MSE, perceptual loss) or, in the decoder-free setting, a next-embedding alignment loss analogous to NE-Dreamer~\citep{anonymous2026nedreamer}.

\paragraph{Latent trajectory regularizer.}
Without intermediate decoding, the latent states at steps $1, \ldots, K-1$ are unsupervised and could drift into regions that are spectrally stable yet semantically meaningless.  We introduce two lightweight constraints:

\begin{enumerate}
  \item \textbf{Latent consistency loss.}
    We encode each intermediate ground-truth observation $o_k$ ($k = 1,
    \ldots, K-1$) with the frozen encoder $\mathcal{E}_\phi$ to obtain
    reference embeddings $e_k^{\star} = \mathrm{sg}(\mathcal{E}_\phi(o_k))$,
    then align:
    \begin{equation}
      \mathcal{L}_{\mathrm{consist}} =
        \frac{1}{K-1}\sum_{k=1}^{K-1}
          \left\|\, g_\omega(h_k) - e_k^{\star}\,\right\|_2^2,
      \label{eq:dd_consist}
    \end{equation}
    where $g_\omega$ is a lightweight projection head and $\mathrm{sg}(\cdot)$
    denotes stop-gradient.  This loss provides soft guidance without
    requiring a full decoder at each step, analogous to the latent
    overshooting technique in PlaNet~\citep{hafner2019planet}.

  \item \textbf{Spectral contraction budget.}
    The spectral-norm constraint on $\bar{A}$ (Section~\ref{sec:dynamics})
    already ensures bounded latent evolution per inner loop.  Over $K$ outer
    steps, we additionally monitor the cumulative contraction:
    \begin{equation}
      \left\|\, h_K - h_0 \,\right\|_2
        \;\leq\; \sum_{k=0}^{K-1} \left\|\, h_{k+1} - h_k \,\right\|_2
        \;\leq\; K \cdot C_{\max},
      \label{eq:dd_contraction}
    \end{equation}
    where $C_{\max}$ is a soft upper bound enforced as a penalty term.
    This prevents latent explosion over long deferred horizons while still
    permitting meaningful state changes induced by actions.
\end{enumerate}

The full training objective for the deferred-decoding variant is:
\begin{equation}
  \mathcal{L}_{\mathrm{DD}} =
    \mathcal{L}_{\mathrm{terminal}}
    + \alpha\,\mathcal{L}_{\mathrm{consist}}
    + \beta\,\max\!\bigl(0,\;\textstyle\sum_{k}\|h_{k+1}-h_k\|_2 - K\cdot C_{\max}\bigr),
  \label{eq:dd_total}
\end{equation}
where $\alpha$ and $\beta$ are balancing coefficients.

\subsubsection{Curriculum over Deferral Horizon $K$}
\label{sec:dd_curriculum}

Training directly with a large $K$ is unstable because gradients must back-propagate through $K \times T$ shared-parameter applications. We adopt a \textbf{progressive horizon curriculum}: training begins with $K = 1$ (equivalent to standard per-step decoding) and gradually increases $K$ during training according to a schedule $K(\text{step}) = \min(K_{\max},\; 1 + \lfloor \text{step} / \Delta \rfloor)$, where $\Delta$ is the number of training steps between increments. This allows the latent dynamics to first learn accurate single-step transitions before being challenged with longer decode-free rollouts.

\subsubsection{Inference Modes}
\label{sec:dd_inference}

Deferred decoding naturally supports two inference modes:

\paragraph{Planning mode.}
Given a candidate action sequence $(a_0, \ldots, a_{K-1})$ from a planner (e.g., CEM, MPPI), the model performs a single decode-free rollout and evaluates only the terminal state $h_K$.  This reduces decoder invocations from $K$ to $1$, saving approximately $(K-1) \times \text{cost}(\mathcal{D}_\xi)$ FLOPs per candidate sequence. When combined with adaptive early exit within each inner loop (Section~\ref{sec:earlyexit}), the total FLOP reduction can reach up to two orders of magnitude for long-horizon planning with simple transitions.

\paragraph{Monitoring mode.}
When intermediate state inspection is needed (e.g., for safety-critical applications), the lightweight projection head $g_\omega$ can be applied at any step $k$ to produce a low-dimensional state summary $\tilde{e}_k = g_\omega(h_k)$ without invoking the full decoder. The full decoder $\mathcal{D}_\xi$ remains available as an optional diagnostic tool but is not required for the planning loop.

\subsubsection{Relationship to Prior Work}
\label{sec:dd_prior}

Table~\ref{tab:dd_comparison} summarizes the key distinctions:

\begin{table}[h]
\centering
\scriptsize
\caption{Comparison of intermediate decoding strategies across world-model
architectures.}
\label{tab:dd_comparison}
\begin{tabular}{lcccc}
\toprule
\textbf{Method} & \textbf{Latent dynamics} & \textbf{Intermediate decode}
  & \textbf{Action injection} & \textbf{Looped depth} \\
\midrule
Dreamer~\citep{hafner2020dreamerv1}
  & RSSM & reward + value at each step & per step & -- \\
MuZero~\citep{schrittwieser2020muzero}
  & learned MLP & policy + value + reward & per step & -- \\
PlaNet~\citep{hafner2019planet}
  & RSSM & reconstruction at each step & per step & -- \\
ETD~\citep{koishekenov2025etd}
  & looped layers & decode only at end & -- (language) & \checkmark \\
NE-Dreamer~\citep{anonymous2026nedreamer}
  & RSSM & embedding alignment & per step & -- \\
\midrule
\textbf{LoopWM-DD (ours)}
  & looped transformer & \textbf{decode only at step $K$}
  & \textbf{per step in latent} & \checkmark \\
\bottomrule
\end{tabular}
\end{table}

Dreamer's imagination rollout already avoids re-encoding real observations but still applies reward and value heads at every imagined step~\citep{hafner2020dreamerv1}. MuZero dispenses with observation reconstruction entirely but uses a non-looped, fixed-depth dynamics function~\citep{schrittwieser2020muzero}. ETD~\citep{koishekenov2025etd} demonstrates the encode-think-decode paradigm for language reasoning with looped layers, but does not handle action-conditioned state transitions or environment simulation. Our deferred decoding unifies these insights: it applies the looped transformer's iterative refinement at each action step in latent space (inheriting the parameter efficiency and spectral stability of the LoopWM) while deferring all observation-space computation to the terminal step, yielding a clean separation between \emph{latent dynamics reasoning} (inner + outer loops) and \emph{observation grounding} (single terminal decode).

\section{Results}
\subsection{Main Results on ScienceWorld}
\begin{table}[h]
\centering
    \setlength\tabcolsep{2.5pt}
    \setlength\extrarowheight{0pt}
\caption{Comparison of our proposed looped world model against claude-opus-4-6-max \citep{anthropic2026opus46} on ScienceWorld dataset \citep{wang-etal-2022-scienceworld} world modelling task. The accuracy is calculated based on feeding consecutive five actions, and obtaining the final scores on world modelling. Note that our model is a model with about 1B model parameters. Refer to Table~\ref{tab:scienceworldmain2} for more baselines.}
\label{tab:scienceworldmain}
\begin{tabular}{lcccc|lcccc}
\toprule
\textbf{Task Type} & \textbf{EM} & \textbf{F1}
  & \textbf{BLEU} & \textbf{Entity} & \textbf{Task Type} & \textbf{EM} & \textbf{F1}
  & \textbf{BLEU} & \textbf{Entity} \\
\midrule
\multicolumn{10}{c}{\textit{Looped World Model (Ours)}}\\
\midrule
Boil  & 66.7\%    &  79.0\%   & 75.3\%    &  77.5\% &
Chemistry        &    44.4\%    &   64.4\% &   54.2\%   &   57.9\%\\
Conductivity       &    87.0\%    &  89.0\%  &  87.8\%    &  87.9\% &
Find               &    76.9\%    &   90.4\%    &82.7\%      &85.8\%\\
Freeze              &    25.0\%    &  59.7\%  &  31.2\%      &54.8\%&
Genetics          &    78.3\%   &   80.2\%   & 78.9\%    &  79.8\%\\
Grow               &    73.8\%   &   80.0\%  &  75.5\%  &    79.8\%&
Incline           &    59.3\%   &   95.3\%  &  90.4\%  &    93.4\%\\
LifeStages          &     0.0\%    &  18.3\%   &  6.1\%      &10.2\%&
Lifespan            &   100.0\%   &  100.0\%  & 100.0\%   &  100.0\%\\
Melt               &    73.0\%    &  91.9\%  &  85.7\%   &   91.6\%&
Power               &    57.1\%   &   63.9\% &   60.8\%    &  61.5\%\\
StateChange         &    80.0\%    &  83.1\%  &  80.0\%   &   80.0\%&
Thermometer         &    83.3\%   &   85.3\%  &  83.3\%    &  83.3\%\\
\midrule
\multicolumn{10}{c}{\textbf{Overall: EM: 68.4\%; Token F1: 85.3\%; BLEU-4: 80.7\%; Entity: 83.9\%}}\\
\midrule
\multicolumn{10}{c}{\textit{claude-opus-4-6-max}}\\
\midrule
Boil                &    22.2\%  &    33.3\%  &  30.2\%    &  32.3\% &
Chemistry          &    44.4\%   &   59.8\%   & 46.3\%    &  59.2\% \\
Conductivity       &    47.8\%  &    67.2\%   & 53.1\%    &  72.7\% &
Find               &    69.2\%   &   83.8\%  &  78.9\%   &   84.6\% \\
Freeze              &    12.5\%  &    33.6\%  &  21.6\%    &  37.3\%&
Genetics          &    59.2\%   &   71.8\%  &  65.7\%    &  71.6\%\\
Grow               &    70.8\%  &    81.6\%  &  76.1\%   &   80.7\%&
Incline           &    34.0\%    &  86.5\%   & 76.2\%    &  83.8\%\\
LifeStages          &     0.0\%  &    10.6\% &    6.0\%   &    6.0\%&
Lifespan            &     0.0\%  &    61.4\%  &   0.0\%    &  58.3\%\\
Melt               &    36.5\%   &   52.4\%   & 46.3\%    &  53.4\%&
Power               &    42.9\%  &    47.3\%  &  45.3\%   &   45.8\%\\
StateChange         &    40.0\%   &   65.5\%  &  44.6\%   &   73.3\%&
Thermometer         &    83.3\%   &   98.1\%  &  93.5\%   &   97.2\%\\
\midrule
\multicolumn{10}{c}{Overall: EM: 47.2\%; Token F1: 72.8\%; BLEU-4: 64.4\%; Entity: 72.3\%}\\
\bottomrule
\end{tabular}
\end{table}

Table~\ref{tab:scienceworldmain} presents the results on the ScienceWorld dataset of our models against claude-opus-4-6-max. From the results, it is clear that our model surpasses the strong claude-opus-4-6-max. In the most extreme cases, it improves the scores on Lifespan from 0\% to 100\%, denoting the underlying strong capacity of our model. On average, our model shows a promising capability, clearly surpassing the baseline by 21.2\% on EM, and clearly on other metrics. Further, we note that our model is a small AI model with around 1B parameters, which is much smaller than those strong closed-source API models such as claude-opus-4-6-max by more than 100x. This suggests our proposed model has a promising parameter efficiency to be deployed on downstream applications. Note that Table~\ref{tab:scienceworldmain2} presents more baselines, which lead to the same conclusions.
\par
We also note that qwen-3.5-flash and gemini-3-flash-preview seem to be clearly worse than other baselines and our models across most metrics. This is reasonable as they are considered smaller than the other baseline models. Our proposed models are still competitive and much stronger than them across the metrics.

\begin{table}[h]
\centering
    \setlength\tabcolsep{2.5pt}
    \setlength\extrarowheight{0pt}
\caption{Baseline results on ScienceWorld dataset \citep{wang-etal-2022-scienceworld} world modelling task. The accuracy is calculated based on feeding consecutive five actions, and obtaining the final scores on world modelling. Note that our model is a model with about 1B model parameters.}
\label{tab:scienceworldmain2}
\begin{tabular}{lcccc|lcccc}
\toprule
\textbf{Task Type} & \textbf{EM} & \textbf{F1}
  & \textbf{BLEU} & \textbf{Entity} & \textbf{Task Type} & \textbf{EM} & \textbf{F1}
  & \textbf{BLEU} & \textbf{Entity} \\
\midrule
\multicolumn{10}{c}{\textit{qwen-3.5-flash \citep{qwen35blog}}}\\
\midrule
Boil                &    0.0\%&      44.6\%&    15.1\%&      39.6\%&
Chemistry          &     3.7\%&      28.8\%&     4.1\%&      41.3\%\\
Conductivity       &     0.0\%&      28.2\%&     0.8\%&      49.2\%&
Find               &     0.0\%&      25.1\%&     0.0\%&      51.0\%\\
Freeze              &     0.0\%&      29.8\%&     9.8\%&      44.0\%&
Genetics          &     7.5\%&      30.5\%&    11.5\%&      53.2\%\\
Grow               &     4.6\%&      28.3\%&     7.2\%&      58.3\%&
Incline           &    20.7\%&      81.3\%&    63.9\%&      84.0\%\\
LifeStages          &     0.0\%&      24.8\%&     0.0\%&      10.2\%&
Lifespan            &     0.0\%&      12.1\%&     0.0\%&      25.0\%\\
Melt               &     9.5\%&      44.1\%&    24.6\%&      64.7\%&
Power               &     0.0\%&      27.5\%&     1.4\%&      38.9\%\\
StateChange         &     0.0\%&      25.1\%&     5.7\%&      70.0\%&
Thermometer         &     0.0\%&      32.7\%&     2.1\%&      56.9\%\\
\midrule
\multicolumn{10}{c}{Overall: EM: 10.0\%; Token F1: 46.9\%; BLEU-4: 26.7\%; Entity: 63.0\%}\\
\midrule
\multicolumn{10}{c}{\textit{gemini-3-flash-preview-thinking \citep{gemini3flash2025}}}\\
\midrule
Boil                &    22.2\%&      61.5\%&    41.9\%&      64.1\%&
Chemistry          &    22.2\%&      54.8\%&    27.6\%&      57.9\%\\
Conductivity       &    17.4\%&      55.4\%&    21.9\%&      60.6\%&
Find               &    15.4\%&      65.2\%&    40.3\%&      80.7\%\\
Freeze              &    12.5\%&      35.0\%&    22.2\%&      37.5\%&
Genetics          &    41.7\%&      65.6\%&    48.5\%&      71.1\%\\
Grow               &    47.7\%&      72.6\%&    55.4\%&      75.8\%&
Incline           &    32.7\%&      88.5\%&    76.0\%&      88.6\%\\
LifeStages          &     0.0\%&      15.1\%&     6.0\%&       8.1\%&
Lifespan            &     0.0\%&      38.8\%&     0.0\%&      58.3\%\\
Melt               &     7.9\%&      47.9\%&    29.6\%&      62.5\%&
Power               &    14.3\%&      42.7\%&    16.7\%&      45.9\%\\
StateChange         &    20.0\%&      57.8\%&    23.9\%&      70.0\%&
Thermometer         &    33.3\%&      68.8\%&    45.3\%&      86.1\%\\
\midrule
\multicolumn{10}{c}{Overall: EM: 30.8\%; Token F1: 68.9\%; BLEU-4: 51.1\%; Entity: 73.8\%}\\
\bottomrule
\end{tabular}
\end{table}
\subsection{Main Results on AlfWorld}
\begin{table}[h]
\centering
    \setlength\tabcolsep{2.5pt}
    \setlength\extrarowheight{0pt}
\caption{Comparison of our proposed looped world model against claude-opus-4-6-max \citep{anthropic2026opus46} and other baselines on AlfWorld dataset \citep{cote2018textworld} world modelling task. The accuracy is calculated based on feeding consecutive five actions, and obtaining the final scores on world modelling. Note that our model is a model with about 1B model parameters.}
\label{tab:alfworldmain}
\begin{tabular}{lcccc|lcccc}
\toprule
\textbf{Task Type} & \textbf{EM} & \textbf{F1}
  & \textbf{BLEU} & \textbf{Entity} & \textbf{Task Type} & \textbf{EM} & \textbf{F1}
  & \textbf{BLEU} & \textbf{Entity} \\
\midrule
\multicolumn{10}{c}{\textit{Looped World Model (Ours)}}\\
\midrule
clean            &    60.4\%&      81.7\%&    75.0\%&      81.3\%&
cool             &    50.0\%&      81.7\%&    72.6\%&      78.5\%\\
heat             &    55.0\%&      81.8\%&    76.6\%&      81.2\%&
look             &    60.5\%&      78.9\%&    73.1\%&      82.0\%\\
pick            &    46.7\%&      79.7\%&    69.1\%&      81.5\%&
- & - & - & - & -\\
\midrule
\multicolumn{10}{c}{\textbf{Overall: EM: 51.6\%; Token F1: 80.4\%; BLEU-4: 71.6\%; Entity: 81.1\%}}\\
\midrule
\multicolumn{10}{c}{\textit{claude-opus-4-6-max}}\\
\midrule
clean            &    57.3\%&      73.8\%&    68.9\%&      77.4\%&
cool             &    50.0\%&      73.0\%&    68.2\%&      72.8\%\\
heat             &    52.5\%&      67.6\%&    64.4\%&      74.1\%&
look             &    60.5\%&      73.2\%&    68.9\%&      78.6\%\\
pick            &    51.0\%&      72.8\%&    65.7\%&      78.1\%&
- & - & - & - & -\\
\midrule
\multicolumn{10}{c}{Overall: EM: 53.0\%; Token F1: 72.6\%; BLEU-4: 66.8\%; Entity: 77.0\%}\\
\midrule
\multicolumn{10}{c}{\textit{qwen-3.5-flash \citep{qwen35blog}}}\\
\midrule
clean            &    36.5\%  &    71.9\% &   55.6\%&      90.1\%&
cool             &    27.3\%   &   72.0\%  &  52.6\%&      85.1\%\\
heat             &    27.5\%   &   70.1\%  &  49.5\%&      91.2\%&
look             &    27.9\%   &   66.2\%  &  46.3\%&      92.8\%\\
pick            &    21.2\%   &   64.1\%  &  43.5\%&      87.5\%&
- & - & - & - & -\\
\midrule
\multicolumn{10}{c}{Overall: EM: 26.0\%; Token F1: 67.3\%; BLEU-4: 47.7\%; Entity: 88.4\%}\\
\midrule
\multicolumn{10}{c}{\textit{gemini-3-flash-preview-thinking \citep{gemini3flash2025}}}\\
\midrule
clean            &    61.5\% &     88.2\%  &  79.9\%&      90.5\%&
cool             &    54.5\% &     88.1\%  &  77.0\%&      88.3\%\\
heat             &    55.0\% &     81.9\%  &  73.3\%&      90.6\%&
look             &    55.8\% &     86.6\%  &  74.4\%&      97.7\%\\
pick            &    42.7\%  &    80.2\%   & 65.1\% &     89.3\%&
- & - & - & - & -\\
\midrule
\multicolumn{10}{c}{Overall: EM: 50.0\%; Token F1: 83.5\%; BLEU-4: 71.0\%; Entity: 90.2\%}\\
\bottomrule
\end{tabular}
\end{table}

Table~\ref{tab:alfworldmain} presents evaluation results on the AlfWorld dataset. On this dataset, we see that the trend can still be promising, as our proposed model gives a promising overall result, given the fact that it is pretty small in terms of model size, with around 1B parameters. Notably, it gives the best result on the BLEU metrics \citep{papineni-etal-2002-bleu} among four models, and ranks in second place on EM and Token F1. Further, by inspecting the detailed action categories, we found that our model seems to have low entity scores, and it seems valid for most action categories. Such an error analysis indicates that future optimization can focus on the entity scores to further enhance the model.

\subsection{Deep Analysis on Deferred Decoding}

\begin{table}[h]
\centering
    \setlength\tabcolsep{15pt}
    \setlength\extrarowheight{0pt}
\caption{The effect of deferred decoding on the ScienceWorld dataset \citep{wang-etal-2022-scienceworld} world modelling task on average over all the tasks, compared to gemini-3-flash-preview-thinking. The relative improvements are calculated using the absolute performance $(Ours - Baselines) / Baselines * 100\%$. Note that our model is a model with about 1B model parameters.}
\label{tab:scienceworld1}
\begin{tabular}{lcccc}
\toprule
\textbf{Task Type} & \textbf{EM} & \textbf{F1}
  & \textbf{BLEU} & \textbf{Entity}\\
\midrule
Step1 & +73.2\% & +16.4\% & +47.0\% & +9.7\% \\
Step2 & +54.5\% & +21.4\% & +41.7\% & +18.0\% \\
Step3 & +103.6\% & +28.1\% & +65.0\% & +19.0\% \\
Step4 & +82.9\% & +29.0\% & +55.5\% & +20.7\% \\
Step5 & +113.8\% & +22.4\% & +54.6\% & +12.8\% \\
\bottomrule
\end{tabular}
\end{table}

\begin{table}[h]
\centering
    \setlength\tabcolsep{15pt}
    \setlength\extrarowheight{0pt}
\caption{The effect of deferred decoding on the ScienceWorld dataset \citep{wang-etal-2022-scienceworld} world modelling task on the task of Boil, compared to gemini-3-flash-preview-thinking. The relative improvements are calculated using the absolute performance $(Ours - Baselines) / Baselines * 100\%$. Note that our model is a model with about 1B model parameters. `—` represents that the baseline score is 0 and LoopWM score is not zero. `0\%` means that both of them has a score of 0.}
\label{tab:scienceworld2}
\begin{tabular}{lcccc}
\toprule
\textbf{Task Type} & \textbf{EM} & \textbf{F1}
  & \textbf{BLEU} & \textbf{Entity}\\
\midrule
Step1 & +100.0\% & +3.5\% & +39.6\% & -8.0\% \\
Step2 & +50.2\% & +33.4\% & +61.3\% & +54.9\% \\
Step3 & +250.5\% & +57.5\% & +136.2\% & +39.3\% \\
Step4 & +700.9\% & +120.0\% & +503.5\% & +121.0\% \\
Step5 & +500.9\% & +29.9\% & +101.9\% & +20.0\% \\
\bottomrule
\end{tabular}
\end{table}

\begin{table}[h]
\centering
    \setlength\tabcolsep{15pt}
    \setlength\extrarowheight{0pt}
\caption{The effect of deferred decoding on the ScienceWorld dataset \citep{wang-etal-2022-scienceworld} world modelling task on the task of Chemistry, compared to gemini-3-flash-preview-thinking. The relative improvements are calculated using the absolute performance $(Ours - Baselines) / Baselines * 100\%$. Note that our model is a model with about 1B model parameters.}
\label{tab:scienceworld3}
\begin{tabular}{lcccc}
\toprule
\textbf{Task Type} & \textbf{EM} & \textbf{F1}
  & \textbf{BLEU} & \textbf{Entity}\\
\midrule
Step1 & +267.1\% & +52.4\% & +197.2\% & +56.9\% \\
Step2 & +140.3\% & +42.2\% & +120.0\% & +33.5\% \\
Step3 & +110.3\% & +34.1\% & +92.5\% & +18.5\% \\
Step4 & +367.6\% & +57.0\% & +224.3\% & +62.6\% \\
Step5 & +100.0\% & +15.0\% & +78.9\% & 0.0\% \\

\bottomrule
\end{tabular}
\end{table}

\begin{table}[h]
\centering
    \setlength\tabcolsep{15pt}
    \setlength\extrarowheight{0pt}
\caption{The effect of deferred decoding on the ScienceWorld dataset \citep{wang-etal-2022-scienceworld} world modelling task on the task of Conductivity, compared to gemini-3-flash-preview-thinking. The relative improvements are calculated using the absolute performance $(Ours - Baselines) / Baselines * 100\%$. Note that our model is a model with about 1B model parameters.}
\label{tab:scienceworld4}
\begin{tabular}{lcccc}
\toprule
\textbf{Task Type} & \textbf{EM} & \textbf{F1}
  & \textbf{BLEU} & \textbf{Entity}\\
\midrule
Step1 & +78.0\% & +17.5\% & +58.8\% & +2.4\% \\
Step2 & +183.1\% & +44.5\% & +190.5\% & +42.5\% \\
Step3 & +220.7\% & +57.5\% & +249.8\% & +39.0\% \\
Step4 & +183.1\% & +53.2\% & +194.8\% & +48.7\% \\
Step5 & +233.3\% & +51.9\% & +218.1\% & +40.0\% \\

\bottomrule
\end{tabular}
\end{table}

\begin{table}[h]
\centering
    \setlength\tabcolsep{15pt}
    \setlength\extrarowheight{0pt}
\caption{The effect of deferred decoding on the ScienceWorld dataset \citep{wang-etal-2022-scienceworld} world modelling task on the task of Find, compared to gemini-3-flash-preview-thinking. The relative improvements are calculated using the absolute performance $(Ours - Baselines) / Baselines * 100\%$. Note that our model is a model with about 1B model parameters.}
\label{tab:scienceworld5}
\begin{tabular}{lcccc}
\toprule
\textbf{Task Type} & \textbf{EM} & \textbf{F1}
  & \textbf{BLEU} & \textbf{Entity}\\
\midrule
Step1 & +166.2\% & +45.0\% & +141.2\% & +28.4\% \\
Step2 & +79.7\% & +7.1\% & +40.6\% & -8.3\% \\
Step3 & +266.2\% & +71.0\% & +253.7\% & +41.7\% \\
Step4 & +71.6\% & +25.6\% & +56.8\% & +14.6\% \\
Step5 & +399.4\% & +38.7\% & +105.2\% & +6.3\% \\

\bottomrule
\end{tabular}
\end{table}

\begin{table}[h]
\centering
    \setlength\tabcolsep{15pt}
    \setlength\extrarowheight{0pt}
\caption{The effect of deferred decoding on the ScienceWorld dataset \citep{wang-etal-2022-scienceworld} world modelling task on the task of Freeze, compared to gemini-3-flash-preview-thinking. The relative improvements are calculated using the absolute performance $(Ours - Baselines) / Baselines * 100\%$. Note that our model is a model with about 1B model parameters.}
\label{tab:scienceworld6}
\begin{tabular}{lcccc}
\toprule
\textbf{Task Type} & \textbf{EM} & \textbf{F1}
  & \textbf{BLEU} & \textbf{Entity}\\
\midrule
Step1 & +100.0\% & -6.2\% & +63.5\% & -32.1\% \\
Step2 & +50.0\% & +10.1\% & +62.2\% & -2.9\% \\
Step3 & +250.0\% & +80.9\% & +112.6\% & +20.6\% \\
Step4 & +400.0\% & +96.7\% & +303.0\% & +76.0\% \\
Step5 & +100.0\% & +70.6\% & +40.5\% & +46.1\% \\

\bottomrule
\end{tabular}
\end{table}

\begin{table}[h]
\centering
    \setlength\tabcolsep{15pt}
    \setlength\extrarowheight{0pt}
\caption{The effect of deferred decoding on the ScienceWorld dataset \citep{wang-etal-2022-scienceworld} world modelling task on the task of Genetics, compared to gemini-3-flash-preview-thinking. The relative improvements are calculated using the absolute performance $(Ours - Baselines) / Baselines * 100\%$. Note that our model is a model with about 1B model parameters.}
\label{tab:scienceworld7}
\begin{tabular}{lcccc}
\toprule
\textbf{Task Type} & \textbf{EM} & \textbf{F1}
  & \textbf{BLEU} & \textbf{Entity}\\
\midrule
Step1 & +80.8\% & +31.9\% & +68.3\% & +27.0\% \\
Step2 & +36.5\% & +24.5\% & +33.0\% & +20.3\% \\
Step3 & +122.1\% & +36.3\% & +101.4\% & +24.3\% \\
Step4 & +76.7\% & +30.5\% & +54.5\% & +22.6\% \\
Step5 & +74.0\% & +19.5\% & +52.3\% & +10.7\% \\

\bottomrule
\end{tabular}
\end{table}

\begin{table}[h]
\centering
    \setlength\tabcolsep{15pt}
    \setlength\extrarowheight{0pt}
\caption{The effect of deferred decoding on the ScienceWorld dataset \citep{wang-etal-2022-scienceworld} world modelling task on the task of Grow, compared to gemini-3-flash-preview-thinking. The relative improvements are calculated using the absolute performance $(Ours - Baselines) / Baselines * 100\%$. Note that our model is a model with about 1B model parameters.}
\label{tab:scienceworld8}
\begin{tabular}{lcccc}
\toprule
\textbf{Task Type} & \textbf{EM} & \textbf{F1}
  & \textbf{BLEU} & \textbf{Entity}\\
\midrule
Step1 & +109.6\% & +20.9\% & +89.1\% & +12.4\% \\
Step2 & +59.6\% & +13.9\% & +46.9\% & +13.9\% \\
Step3 & +48.4\% & +7.6\% & +41.1\% & +6.8\% \\
Step4 & +16.3\% & -5.6\% & +5.8\% & -10.4\% \\
Step5 & +50.0\% & +8.4\% & +34.1\% & +2.8\% \\

\bottomrule
\end{tabular}
\end{table}

\begin{table}[h]
\centering
    \setlength\tabcolsep{15pt}
    \setlength\extrarowheight{0pt}
\caption{The effect of deferred decoding on the ScienceWorld dataset \citep{wang-etal-2022-scienceworld} world modelling task on the task of Incline, compared to gemini-3-flash-preview-thinking. The relative improvements are calculated using the absolute performance $(Ours - Baselines) / Baselines * 100\%$. Note that our model is a model with about 1B model parameters.}
\label{tab:scienceworld9}
\begin{tabular}{lcccc}
\toprule
\textbf{Task Type} & \textbf{EM} & \textbf{F1}
  & \textbf{BLEU} & \textbf{Entity}\\
\midrule
Step1 & +24.5\% & +1.9\% & +7.5\% & -0.1\% \\
Step2 & +7.6\% & +2.3\% & +5.5\% & +2.1\% \\
Step3 & +42.5\% & +6.6\% & +13.9\% & +3.8\% \\
Step4 & +40.2\% & +6.8\% & +14.8\% & +3.3\% \\
Step5 & +85.3\% & +8.4\% & +20.5\% & +6.1\% \\

\bottomrule
\end{tabular}
\end{table}

\begin{table}[h]
\centering
    \setlength\tabcolsep{15pt}
    \setlength\extrarowheight{0pt}
\caption{The effect of deferred decoding on the ScienceWorld dataset \citep{wang-etal-2022-scienceworld} world modelling task on the task of Melt, compared to gemini-3-flash-preview-thinking. The relative improvements are calculated using the absolute performance $(Ours - Baselines) / Baselines * 100\%$. Note that our model is a model with about 1B model parameters.}
\label{tab:scienceworld10}
\begin{tabular}{lcccc}
\toprule
\textbf{Task Type} & \textbf{EM} & \textbf{F1}
  & \textbf{BLEU} & \textbf{Entity}\\
\midrule
Step1 & +112.6\% & -5.7\% & +27.9\% & -22.9\% \\
Step2 & +343.2\% & +105.4\% & +201.4\% & +88.0\% \\
Step3 & +349.6\% & +86.0\% & +172.7\% & +59.4\% \\
Step4 & +585.3\% & +220.3\% & +467.1\% & +138.3\% \\
Step5 & +557.7\% & +79.5\% & +161.3\% & +42.9\% \\

\bottomrule
\end{tabular}
\end{table}

\begin{table}[h]
\centering
    \setlength\tabcolsep{15pt}
    \setlength\extrarowheight{0pt}
\caption{The effect of deferred decoding on the ScienceWorld dataset \citep{wang-etal-2022-scienceworld} world modelling task on the task of Power, compared to gemini-3-flash-preview-thinking. The relative improvements are calculated using the absolute performance $(Ours - Baselines) / Baselines * 100\%$. Note that our model is a model with about 1B model parameters.}
\label{tab:scienceworld11}
\begin{tabular}{lcccc}
\toprule
\textbf{Task Type} & \textbf{EM} & \textbf{F1}
  & \textbf{BLEU} & \textbf{Entity}\\
\midrule
Step1 & +499.3\% & +105.5\% & +499.3\% & +61.6\% \\
Step2 & +99.8\% & +30.6\% & +80.9\% & +17.1\% \\
Step3 & +299.3\% & +61.0\% & +347.5\% & +24.5\% \\
Step4 & +66.4\% & +12.4\% & +66.4\% & -9.8\% \\
Step5 & +299.3\% & +46.2\% & +264.1\% & +34.0\% \\

\bottomrule
\end{tabular}
\end{table}

\begin{table}[h]
\centering
    \setlength\tabcolsep{15pt}
    \setlength\extrarowheight{0pt}
\caption{The effect of deferred decoding on the ScienceWorld dataset \citep{wang-etal-2022-scienceworld} world modelling task averaged on all tasks, compared to qwen3.5-flash. The relative improvements are calculated using the absolute performance $(Ours - Baselines) / Baselines * 100\%$. Note that our model is a model with about 1B model parameters.}
\label{tab:scienceworld12}
\begin{tabular}{lcccc}
\toprule
\textbf{Task Type} & \textbf{EM} & \textbf{F1}
  & \textbf{BLEU} & \textbf{Entity}\\
\midrule
Step1 & +143.5\% & +32.4\% & +93.8\% & +26.3\% \\
Step2 & +86.4\% & +33.9\% & +72.5\% & +27.8\% \\
Step3 & +136.1\% & +42.0\% & +106.0\% & +30.2\% \\
Step4 & +78.1\% & +35.5\% & +74.0\% & +25.1\% \\
Step5 & +104.8\% & +32.2\% & +79.3\% & +22.1\% \\

\bottomrule
\end{tabular}
\end{table}

\begin{table}[h]
\centering
    \setlength\tabcolsep{15pt}
    \setlength\extrarowheight{0pt}
\caption{The effect of deferred decoding on the ScienceWorld dataset \citep{wang-etal-2022-scienceworld} world modelling task averaged on Boil, compared to qwen3.5-flash. The relative improvements are calculated using the absolute performance $(Ours - Baselines) / Baselines * 100\%$. Note that our model is a model with about 1B model parameters.}
\label{tab:scienceworld13}
\begin{tabular}{lcccc}
\toprule
\textbf{Task Type} & \textbf{EM} & \textbf{F1}
  & \textbf{BLEU} & \textbf{Entity}\\
\midrule
Step1 & — & +38.6\% & +356.5\% & +9.2\% \\
Step2 & — & +116.7\% & +2428.1\% & +140.3\% \\
Step3 & — & +100.2\% & +438.1\% & +55.7\% \\
Step4 & — & +203.1\% & — & +194.7\% \\
Step5 & +500.9\% & +57.1\% & +184.2\% & +86.7\% \\

\bottomrule
\end{tabular}
\end{table}

\begin{table}[h]
\centering
    \setlength\tabcolsep{15pt}
    \setlength\extrarowheight{0pt}
\caption{The effect of deferred decoding on the ScienceWorld dataset \citep{wang-etal-2022-scienceworld} world modelling task averaged on Chemistry, compared to qwen3.5-flash. The relative improvements are calculated using the absolute performance $(Ours - Baselines) / Baselines * 100\%$. Note that our model is a model with about 1B model parameters.}
\label{tab:scienceworld14}
\begin{tabular}{lcccc}
\toprule
\textbf{Task Type} & \textbf{EM} & \textbf{F1}
  & \textbf{BLEU} & \textbf{Entity}\\
\midrule
Step1 & +450.7\% & +114.6\% & +369.8\% & +103.3\% \\
Step2 & +500.7\% & +111.4\% & +487.7\% & +86.4\% \\
Step3 & +600.9\% & +87.3\% & +411.0\% & +38.1\% \\
Step4 & +250.7\% & +61.6\% & +217.6\% & +43.0\% \\
Step5 & +300.9\% & +43.4\% & +290.6\% & +15.3\% \\

\bottomrule
\end{tabular}
\end{table}

\begin{table}[h]
\centering
    \setlength\tabcolsep{15pt}
    \setlength\extrarowheight{0pt}
\caption{The effect of deferred decoding on the ScienceWorld dataset \citep{wang-etal-2022-scienceworld} world modelling task averaged on Conductivity, compared to qwen3.5-flash. The relative improvements are calculated using the absolute performance $(Ours - Baselines) / Baselines * 100\%$. Note that our model is a model with about 1B model parameters.}
\label{tab:scienceworld15}
\begin{tabular}{lcccc}
\toprule
\textbf{Task Type} & \textbf{EM} & \textbf{F1}
  & \textbf{BLEU} & \textbf{Entity}\\
\midrule
Step1 & +100.0\% & +19.3\% & +71.3\% & -2.4\% \\
Step2 & +240.5\% & +40.8\% & +209.7\% & +26.0\% \\
Step3 & +166.3\% & +27.0\% & +169.3\% & +1.3\% \\
Step4 & +143.1\% & +45.3\% & +191.0\% & +23.9\% \\
Step5 & +233.3\% & +39.6\% & +191.7\% & +10.6\% \\

\bottomrule
\end{tabular}
\end{table}

\begin{table}[h]
\centering
    \setlength\tabcolsep{15pt}
    \setlength\extrarowheight{0pt}
\caption{The effect of deferred decoding on the ScienceWorld dataset \citep{wang-etal-2022-scienceworld} world modelling task averaged on Find, compared to qwen3.5-flash. The relative improvements are calculated using the absolute performance $(Ours - Baselines) / Baselines * 100\%$. Note that our model is a model with about 1B model parameters.}
\label{tab:scienceworld16}
\begin{tabular}{lcccc}
\toprule
\textbf{Task Type} & \textbf{EM} & \textbf{F1}
  & \textbf{BLEU} & \textbf{Entity}\\
\midrule
Step1 & +699.2\% & +75.0\% & +400.0\% & +58.9\% \\
Step2 & +349.4\% & +38.6\% & +196.7\% & -0.3\% \\
Step3 & +449.4\% & +105.8\% & +440.5\% & +63.1\% \\
Step4 & +139.7\% & +54.0\% & +133.3\% & +30.0\% \\
Step5 & +149.4\% & +50.7\% & +108.8\% & +8.1\% \\

\bottomrule
\end{tabular}
\end{table}

\begin{table}[h]
\centering
    \setlength\tabcolsep{15pt}
    \setlength\extrarowheight{0pt}
\caption{The effect of deferred decoding on the ScienceWorld dataset \citep{wang-etal-2022-scienceworld} world modelling task averaged on Freeze, compared to qwen3.5-flash. The relative improvements are calculated using the absolute performance $(Ours - Baselines) / Baselines * 100\%$. Note that our model is a model with about 1B model parameters.}
\label{tab:scienceworld17}
\begin{tabular}{lcccc}
\toprule
\textbf{Task Type} & \textbf{EM} & \textbf{F1}
  & \textbf{BLEU} & \textbf{Entity}\\
\midrule
Step1 & +100.0\% & -6.3\% & +308.7\% & -31.3\% \\
Step2 & +200.0\% & +47.7\% & +116.6\% & +33.4\% \\
Step3 & — & +162.2\% & +750.0\% & +51.5\% \\
Step4 & +400.0\% & +109.2\% & +219.7\% & +59.9\% \\
Step5 & — & +63.1\% & +235.5\% & +14.9\% \\

\bottomrule
\end{tabular}
\end{table}

\begin{table}[h]
\centering
    \setlength\tabcolsep{15pt}
    \setlength\extrarowheight{0pt}
\caption{The effect of deferred decoding on the ScienceWorld dataset \citep{wang-etal-2022-scienceworld} world modelling task averaged on Genetics, compared to qwen3.5-flash. The relative improvements are calculated using the absolute performance $(Ours - Baselines) / Baselines * 100\%$. Note that our model is a model with about 1B model parameters.}
\label{tab:scienceworld18}
\begin{tabular}{lcccc}
\toprule
\textbf{Task Type} & \textbf{EM} & \textbf{F1}
  & \textbf{BLEU} & \textbf{Entity}\\
\midrule
Step1 & +203.5\% & +47.2\% & +184.8\% & +53.5\% \\
Step2 & +61.6\% & +23.0\% & +50.6\% & +23.2\% \\
Step3 & +202.9\% & +44.6\% & +185.0\% & +39.9\% \\
Step4 & +70.8\% & +25.8\% & +59.8\% & +22.6\% \\
Step5 & +104.4\% & +28.5\% & +91.5\% & +21.1\% \\

\bottomrule
\end{tabular}
\end{table}

\begin{table}[h]
\centering
    \setlength\tabcolsep{15pt}
    \setlength\extrarowheight{0pt}
\caption{The effect of deferred decoding on the ScienceWorld dataset \citep{wang-etal-2022-scienceworld} world modelling task averaged on Grow, compared to qwen3.5-flash. The relative improvements are calculated using the absolute performance $(Ours - Baselines) / Baselines * 100\%$. Note that our model is a model with about 1B model parameters.}
\label{tab:scienceworld19}
\begin{tabular}{lcccc}
\toprule
\textbf{Task Type} & \textbf{EM} & \textbf{F1}
  & \textbf{BLEU} & \textbf{Entity}\\
\midrule
Step1 & +450.4\% & +63.1\% & +376.8\% & +45.1\% \\
Step2 & +168.8\% & +46.5\% & +135.0\% & +34.6\% \\
Step3 & +257.8\% & +45.7\% & +206.3\% & +31.5\% \\
Step4 & +87.0\% & +15.4\% & +73.1\% & +0.8\% \\
Step5 & +152.7\% & +33.3\% & +125.4\% & +20.4\% \\

\bottomrule
\end{tabular}
\end{table}

\begin{table}[h]
\centering
    \setlength\tabcolsep{15pt}
    \setlength\extrarowheight{0pt}
\caption{The effect of deferred decoding on the ScienceWorld dataset \citep{wang-etal-2022-scienceworld} world modelling task averaged on Incline, compared to qwen3.5-flash. The relative improvements are calculated using the absolute performance $(Ours - Baselines) / Baselines * 100\%$. Note that our model is a model with about 1B model parameters.}
\label{tab:scienceworld20}
\begin{tabular}{lcccc}
\toprule
\textbf{Task Type} & \textbf{EM} & \textbf{F1}
  & \textbf{BLEU} & \textbf{Entity}\\
\midrule
Step1 & +40.2\% & +5.5\% & +12.9\% & +4.1\% \\
Step2 & -1.0\% & +5.8\% & +8.4\% & +4.5\% \\
Step3 & +7.4\% & +8.2\% & +13.5\% & +5.0\% \\
Step4 & +2.4\% & +9.5\% & +13.6\% & +6.4\% \\
Step5 & +9.8\% & +7.1\% & +13.0\% & +4.8\% \\

\bottomrule
\end{tabular}
\end{table}

\begin{table}[h]
\centering
    \setlength\tabcolsep{15pt}
    \setlength\extrarowheight{0pt}
\caption{The effect of deferred decoding on the ScienceWorld dataset \citep{wang-etal-2022-scienceworld} world modelling task averaged on Melt, compared to qwen3.5-flash. The relative improvements are calculated using the absolute performance $(Ours - Baselines) / Baselines * 100\%$. Note that our model is a model with about 1B model parameters.}
\label{tab:scienceworld21}
\begin{tabular}{lcccc}
\toprule
\textbf{Task Type} & \textbf{EM} & \textbf{F1}
  & \textbf{BLEU} & \textbf{Entity}\\
\midrule
Step1 & +241.8\% & +25.0\% & +132.4\% & +5.6\% \\
Step2 & +680.9\% & +143.0\% & +418.9\% & +125.0\% \\
Step3 & +501.1\% & +152.9\% & +368.9\% & +115.7\% \\
Step4 & +723.4\% & +198.4\% & +555.0\% & +150.8\% \\
Step5 & +822.8\% & +128.0\% & +365.8\% & +112.0\% \\

\bottomrule
\end{tabular}
\end{table}

\begin{table}[h]
\centering
    \setlength\tabcolsep{15pt}
    \setlength\extrarowheight{0pt}
\caption{The effect of deferred decoding on the ScienceWorld dataset \citep{wang-etal-2022-scienceworld} world modelling task averaged on Power, compared to qwen3.5-flash. The relative improvements are calculated using the absolute performance $(Ours - Baselines) / Baselines * 100\%$. Note that our model is a model with about 1B model parameters.}
\label{tab:scienceworld22}
\begin{tabular}{lcccc}
\toprule
\textbf{Task Type} & \textbf{EM} & \textbf{F1}
  & \textbf{BLEU} & \textbf{Entity}\\
\midrule
Step1 & +499.3\% & +138.0\% & +350.9\% & +121.2\% \\
Step2 & +499.3\% & +116.3\% & +439.4\% & +46.9\% \\
Step3 & +299.3\% & +77.9\% & +300.6\% & +70.3\% \\
Step4 & — & +103.1\% & +891.7\% & +13.0\% \\
Step5 & +299.3\% & +76.0\% & +272.1\% & +48.2\% \\

\bottomrule
\end{tabular}
\end{table}

\begin{table}[h]
\centering
    \setlength\tabcolsep{15pt}
    \setlength\extrarowheight{0pt}
\caption{The effect of deferred decoding on the ScienceWorld dataset \citep{wang-etal-2022-scienceworld} world modelling task averaged, on our model. Note that our model is a model with about 1B model parameters.}
\label{tab:scienceworld23}
\begin{tabular}{lcccc}
\toprule
\textbf{Task Type} & \textbf{EM} & \textbf{F1}
  & \textbf{BLEU} & \textbf{Entity}\\
\midrule
Step 1 & 67.2\% & 78.0\% & 72.3\% & 77.9\% \\
Step 2 & 68.6\% & 86.2\% & 80.9\% & 86.4\% \\
Step 3 & 68.0\% & 87.5\% & 82.0\% & 87.1\% \\
Step 4 & 68.4\% & 87.1\% & 82.1\% & 85.6\% \\
Step 5 & 68.4\% & 85.3\% & 80.7\% & 83.9\% \\

\bottomrule
\end{tabular}
\end{table}

\begin{table}[h]
\centering
    \setlength\tabcolsep{15pt}
    \setlength\extrarowheight{0pt}
\caption{The effect of deferred decoding on the ScienceWorld dataset \citep{wang-etal-2022-scienceworld} world modelling tasks, on Step 1, on our model. Note that our model is a model with about 1B model parameters.}
\label{tab:scienceworld24}
\begin{tabular}{lcccc}
\toprule
\textbf{Task Type} & \textbf{EM} & \textbf{F1}
  & \textbf{BLEU} & \textbf{Entity}\\
\midrule
Boil & 44.4\% & 59.6\% & 52.5\% & 54.4\% \\
Chemistry & 81.5\% & 85.2\% & 84.1\% & 85.2\% \\
Conductivity & 69.6\% & 78.5\% & 71.6\% & 76.6\% \\
Find & 61.5\% & 73.5\% & 61.5\% & 76.9\% \\
Freeze & 25.0\% & 42.0\% & 34.0\% & 42.5\% \\
Genetics & 78.3\% & 83.9\% & 78.6\% & 85.5\% \\
Grow & 67.7\% & 75.7\% & 67.7\% & 76.3\% \\
Incline & 74.7\% & 91.9\% & 87.7\% & 90.9\% \\
Melt & 27.0\% & 39.5\% & 31.6\% & 38.0\% \\
Power & 85.7\% & 97.6\% & 85.7\% & 100.0\% \\

\bottomrule
\end{tabular}
\end{table}

\begin{table}[h]
\centering
    \setlength\tabcolsep{15pt}
    \setlength\extrarowheight{0pt}
\caption{The effect of deferred decoding on the ScienceWorld dataset \citep{wang-etal-2022-scienceworld} world modelling tasks, on Step 2, on our model. Note that our model is a model with about 1B model parameters.}
\label{tab:scienceworld25}
\begin{tabular}{lcccc}
\toprule
\textbf{Task Type} & \textbf{EM} & \textbf{F1}
  & \textbf{BLEU} & \textbf{Entity}\\
\midrule
Boil & 66.7\% & 81.9\% & 71.6\% & 88.9\% \\
Chemistry & 88.9\% & 92.6\% & 91.1\% & 93.2\% \\
Conductivity & 73.9\% & 83.5\% & 82.2\% & 81.8\% \\
Find & 69.2\% & 78.3\% & 70.3\% & 70.3\% \\
Freeze & 37.5\% & 69.7\% & 54.0\% & 66.7\% \\
Genetics & 80.8\% & 84.9\% & 82.2\% & 85.9\% \\
Grow & 78.5\% & 84.4\% & 79.9\% & 87.1\% \\
Incline & 56.7\% & 93.4\% & 86.9\% & 92.0\% \\
Melt & 49.2\% & 72.9\% & 63.3\% & 73.9\% \\
Power & 85.7\% & 95.6\% & 89.0\% & 98.0\% \\

\bottomrule
\end{tabular}
\end{table}

\begin{table}[h]
\centering
    \setlength\tabcolsep{15pt}
    \setlength\extrarowheight{0pt}
\caption{The effect of deferred decoding on the ScienceWorld dataset \citep{wang-etal-2022-scienceworld} world modelling tasks, on Step 3, on our model. Note that our model is a model with about 1B model parameters.}
\label{tab:scienceworld26}
\begin{tabular}{lcccc}
\toprule
\textbf{Task Type} & \textbf{EM} & \textbf{F1}
  & \textbf{BLEU} & \textbf{Entity}\\
\midrule
Boil & 77.8\% & 97.5\% & 94.7\% & 98.1\% \\
Chemistry & 77.8\% & 85.4\% & 80.1\% & 83.8\% \\
Conductivity & 69.6\% & 82.7\% & 80.8\% & 80.6\% \\
Find & 84.6\% & 96.1\% & 90.9\% & 96.2\% \\
Freeze & 87.5\% & 98.6\% & 70.8\% & 97.9\% \\
Genetics & 83.3\% & 86.3\% & 83.8\% & 85.0\% \\
Grow & 66.2\% & 73.3\% & 68.3\% & 75.1\% \\
Incline & 58.0\% & 96.6\% & 91.2\% & 94.6\% \\
Melt & 57.1\% & 88.0\% & 76.9\% & 92.3\% \\
Power & 57.1\% & 74.7\% & 72.5\% & 72.2\% \\

\bottomrule
\end{tabular}
\end{table}

\begin{table}[h]
\centering
    \setlength\tabcolsep{15pt}
    \setlength\extrarowheight{0pt}
\caption{The effect of deferred decoding on the ScienceWorld dataset \citep{wang-etal-2022-scienceworld} world modelling tasks, on Step 4, on our model. Note that our model is a model with about 1B model parameters.}
\label{tab:scienceworld27}
\begin{tabular}{lcccc}
\toprule
\textbf{Task Type} & \textbf{EM} & \textbf{F1}
  & \textbf{BLEU} & \textbf{Entity}\\
\midrule
Boil & 88.9\% & 98.8\% & 85.1\% & 98.1\% \\
Chemistry & 51.9\% & 72.7\% & 61.3\% & 66.5\% \\
Conductivity & 73.9\% & 93.0\% & 90.2\% & 92.5\% \\
Find & 92.3\% & 94.1\% & 93.3\% & 93.4\% \\
Freeze & 62.5\% & 95.4\% & 66.5\% & 91.7\% \\
Genetics & 82.5\% & 84.3\% & 83.1\% & 84.6\% \\
Grow & 66.2\% & 72.1\% & 67.7\% & 71.3\% \\
Incline & 60.7\% & 95.8\% & 90.9\% & 93.7\% \\
Melt & 65.1\% & 91.6\% & 84.5\% & 90.3\% \\
Power & 71.4\% & 79.0\% & 71.4\% & 71.4\% \\

\bottomrule
\end{tabular}
\end{table}

\begin{table}[h]
\centering
    \setlength\tabcolsep{15pt}
    \setlength\extrarowheight{0pt}
\caption{The effect of deferred decoding on the ScienceWorld dataset \citep{wang-etal-2022-scienceworld} world modelling tasks, on Step 5, on our model. Note that our model is a model with about 1B model parameters.}
\label{tab:scienceworld28}
\begin{tabular}{lcccc}
\toprule
\textbf{Task Type} & \textbf{EM} & \textbf{F1}
  & \textbf{BLEU} & \textbf{Entity}\\
\midrule
Boil & 66.7\% & 79.0\% & 75.3\% & 77.5\% \\
Chemistry & 44.4\% & 64.4\% & 54.2\% & 57.9\% \\
Conductivity & 87.0\% & 89.0\% & 87.8\% & 87.9\% \\
Find & 76.9\% & 90.4\% & 82.7\% & 85.8\% \\
Freeze & 25.0\% & 59.7\% & 31.2\% & 54.8\% \\
Genetics & 78.3\% & 80.2\% & 78.9\% & 79.8\% \\
Grow & 73.8\% & 80.0\% & 75.5\% & 79.8\% \\
Incline & 59.3\% & 95.3\% & 90.4\% & 93.4\% \\
Melt & 73.0\% & 91.9\% & 85.7\% & 91.6\% \\
Power & 57.1\% & 63.9\% & 60.8\% & 61.5\% \\

\bottomrule
\end{tabular}
\end{table}

\begin{table}[h]
\centering
    \setlength\tabcolsep{15pt}
    \setlength\extrarowheight{0pt}
\caption{The effect of deferred decoding on the ScienceWorld dataset \citep{wang-etal-2022-scienceworld} world modelling tasks, on gemini.}
\label{tab:scienceworld29}
\begin{tabular}{lcccc}
\toprule
\textbf{Task Type} & \textbf{EM} & \textbf{F1}
  & \textbf{BLEU} & \textbf{Entity}\\
\midrule
Step 1 & 38.8\% & 67.0\% & 49.2\% & 71.0\% \\
Step 2 & 44.4\% & 71.0\% & 57.1\% & 73.2\% \\
Step 3 & 33.4\% & 68.3\% & 49.7\% & 73.2\% \\
Step 4 & 37.4\% & 67.5\% & 52.8\% & 70.9\% \\
Step 5 & 32.0\% & 69.7\% & 52.2\% & 74.4\% \\

\bottomrule
\end{tabular}
\end{table}

\begin{table}[h]
\centering
    \setlength\tabcolsep{15pt}
    \setlength\extrarowheight{0pt}
\caption{The effect of deferred decoding on the ScienceWorld dataset \citep{wang-etal-2022-scienceworld} world modelling tasks, on Step 1 on gemini.}
\label{tab:scienceworld30}
\begin{tabular}{lcccc}
\toprule
\textbf{Task Type} & \textbf{EM} & \textbf{F1}
  & \textbf{BLEU} & \textbf{Entity}\\
\midrule
Boil & 22.2\% & 57.6\% & 37.6\% & 59.1\% \\
Chemistry & 22.2\% & 55.9\% & 28.3\% & 54.3\% \\
Conductivity & 39.1\% & 66.8\% & 45.1\% & 74.8\% \\
Find & 23.1\% & 50.7\% & 25.5\% & 59.9\% \\
Freeze & 12.5\% & 44.8\% & 20.8\% & 62.6\% \\
Genetics & 43.3\% & 63.6\% & 46.7\% & 67.3\% \\
Grow & 32.3\% & 62.6\% & 35.8\% & 67.9\% \\
Incline & 60.0\% & 90.2\% & 81.6\% & 91.0\% \\
Melt & 12.7\% & 41.9\% & 24.7\% & 49.3\% \\
Power & 14.3\% & 47.5\% & 14.3\% & 61.9\% \\

\bottomrule
\end{tabular}
\end{table}

\begin{table}[h]
\centering
    \setlength\tabcolsep{15pt}
    \setlength\extrarowheight{0pt}
\caption{The effect of deferred decoding on the ScienceWorld dataset \citep{wang-etal-2022-scienceworld} world modelling tasks, on Step 2 on gemini.}
\label{tab:scienceworld31}
\begin{tabular}{lcccc}
\toprule
\textbf{Task Type} & \textbf{EM} & \textbf{F1}
  & \textbf{BLEU} & \textbf{Entity}\\
\midrule
Boil & 44.4\% & 61.4\% & 44.4\% & 57.4\% \\
Chemistry & 37.0\% & 65.1\% & 41.4\% & 69.8\% \\
Conductivity & 26.1\% & 57.8\% & 28.3\% & 57.4\% \\
Find & 38.5\% & 73.1\% & 50.0\% & 76.7\% \\
Freeze & 25.0\% & 63.3\% & 33.3\% & 68.7\% \\
Genetics & 59.2\% & 68.2\% & 61.8\% & 71.4\% \\
Grow & 49.2\% & 74.1\% & 54.4\% & 76.5\% \\
Incline & 52.7\% & 91.3\% & 82.4\% & 90.1\% \\
Melt & 11.1\% & 35.5\% & 21.0\% & 39.3\% \\
Power & 42.9\% & 73.2\% & 49.2\% & 83.7\% \\

\bottomrule
\end{tabular}
\end{table}

\begin{table}[h]
\centering
    \setlength\tabcolsep{15pt}
    \setlength\extrarowheight{0pt}
\caption{The effect of deferred decoding on the ScienceWorld dataset \citep{wang-etal-2022-scienceworld} world modelling tasks, on Step 3 on gemini.}
\label{tab:scienceworld32}
\begin{tabular}{lcccc}
\toprule
\textbf{Task Type} & \textbf{EM} & \textbf{F1}
  & \textbf{BLEU} & \textbf{Entity}\\
\midrule
Boil & 22.2\% & 61.9\% & 40.1\% & 70.4\% \\
Chemistry & 37.0\% & 63.7\% & 41.6\% & 70.7\% \\
Conductivity & 21.7\% & 52.5\% & 23.1\% & 58.0\% \\
Find & 23.1\% & 56.2\% & 25.7\% & 67.9\% \\
Freeze & 25.0\% & 54.5\% & 33.3\% & 81.2\% \\
Genetics & 37.5\% & 63.3\% & 41.6\% & 68.4\% \\
Grow & 44.6\% & 68.1\% & 48.4\% & 70.3\% \\
Incline & 40.7\% & 90.6\% & 80.1\% & 91.1\% \\
Melt & 12.7\% & 47.3\% & 28.2\% & 57.9\% \\
Power & 14.3\% & 46.4\% & 16.2\% & 58.0\% \\

\bottomrule
\end{tabular}
\end{table}

\begin{table}[h]
\centering
    \setlength\tabcolsep{15pt}
    \setlength\extrarowheight{0pt}
\caption{The effect of deferred decoding on the ScienceWorld dataset \citep{wang-etal-2022-scienceworld} world modelling tasks, on Step 4 on gemini.}
\label{tab:scienceworld33}
\begin{tabular}{lcccc}
\toprule
\textbf{Task Type} & \textbf{EM} & \textbf{F1}
  & \textbf{BLEU} & \textbf{Entity}\\
\midrule
Boil & 11.1\% & 44.9\% & 14.1\% & 44.4\% \\
Chemistry & 11.1\% & 46.3\% & 18.9\% & 40.9\% \\
Conductivity & 26.1\% & 60.7\% & 30.6\% & 62.2\% \\
Find & 53.8\% & 74.9\% & 59.5\% & 81.5\% \\
Freeze & 12.5\% & 48.5\% & 16.5\% & 52.1\% \\
Genetics & 46.7\% & 64.6\% & 53.8\% & 69.0\% \\
Grow & 56.9\% & 76.4\% & 64.0\% & 79.6\% \\
Incline & 43.3\% & 89.7\% & 79.2\% & 90.7\% \\
Melt & 9.5\% & 28.6\% & 14.9\% & 37.9\% \\
Power & 42.9\% & 70.3\% & 42.9\% & 79.2\% \\

\bottomrule
\end{tabular}
\end{table}

\begin{table}[h]
\centering
    \setlength\tabcolsep{15pt}
    \setlength\extrarowheight{0pt}
\caption{The effect of deferred decoding on the ScienceWorld dataset \citep{wang-etal-2022-scienceworld} world modelling tasks, on Step 5 on gemini.}
\label{tab:scienceworld34}
\begin{tabular}{lcccc}
\toprule
\textbf{Task Type} & \textbf{EM} & \textbf{F1}
  & \textbf{BLEU} & \textbf{Entity}\\
\midrule
Boil & 11.1\% & 60.8\% & 37.3\% & 64.6\% \\
Chemistry & 22.2\% & 56.0\% & 30.3\% & 57.9\% \\
Conductivity & 26.1\% & 58.6\% & 27.6\% & 62.8\% \\
Find & 15.4\% & 65.2\% & 40.3\% & 80.7\% \\
Freeze & 12.5\% & 35.0\% & 22.2\% & 37.5\% \\
Genetics & 45.0\% & 67.1\% & 51.8\% & 72.1\% \\
Grow & 49.2\% & 73.8\% & 56.3\% & 77.6\% \\
Incline & 32.0\% & 87.9\% & 75.0\% & 88.0\% \\
Melt & 11.1\% & 51.2\% & 32.8\% & 64.1\% \\
Power & 14.3\% & 43.7\% & 16.7\% & 45.9\% \\

\bottomrule
\end{tabular}
\end{table}

\begin{table}[h]
\centering
    \setlength\tabcolsep{15pt}
    \setlength\extrarowheight{0pt}
\caption{The effect of deferred decoding on the ScienceWorld dataset \citep{wang-etal-2022-scienceworld} world modelling tasks on qwen.}
\label{tab:scienceworld35}
\begin{tabular}{lcccc}
\toprule
\textbf{Task Type} & \textbf{EM} & \textbf{F1}
  & \textbf{BLEU} & \textbf{Entity}\\
\midrule
Step 1 & 27.6\% & 58.9\% & 37.3\% & 61.7\% \\
Step 2 & 36.8\% & 64.4\% & 46.9\% & 67.6\% \\
Step 3 & 28.8\% & 61.6\% & 39.8\% & 66.9\% \\
Step 4 & 38.4\% & 64.3\% & 47.2\% & 68.4\% \\
Step 5 & 33.4\% & 64.5\% & 45.0\% & 68.7\% \\

\bottomrule
\end{tabular}
\end{table}

\begin{table}[h]
\centering
    \setlength\tabcolsep{15pt}
    \setlength\extrarowheight{0pt}
\caption{The effect of deferred decoding on the ScienceWorld dataset \citep{wang-etal-2022-scienceworld} world modelling tasks, on Step 1 on qwen.}
\label{tab:scienceworld36}
\begin{tabular}{lcccc}
\toprule
\textbf{Task Type} & \textbf{EM} & \textbf{F1}
  & \textbf{BLEU} & \textbf{Entity}\\
\midrule
Boil & 0.0\% & 43.0\% & 11.5\% & 49.8\% \\
Chemistry & 14.8\% & 39.7\% & 17.9\% & 41.9\% \\
Conductivity & 34.8\% & 65.8\% & 41.8\% & 78.5\% \\
Find & 7.7\% & 42.0\% & 12.3\% & 48.4\% \\
Freeze & 0.0\% & 34.3\% & 8.3\% & 49.3\% \\
Genetics & 25.8\% & 57.0\% & 27.6\% & 55.7\% \\
Grow & 12.3\% & 46.4\% & 14.2\% & 52.6\% \\
Incline & 53.3\% & 87.1\% & 77.7\% & 87.3\% \\
Melt & 7.9\% & 31.6\% & 13.6\% & 36.0\% \\
Power & 14.3\% & 41.0\% & 19.0\% & 45.2\% \\

\bottomrule
\end{tabular}
\end{table}

\begin{table}[h]
\centering
    \setlength\tabcolsep{15pt}
    \setlength\extrarowheight{0pt}
\caption{The effect of deferred decoding on the ScienceWorld dataset \citep{wang-etal-2022-scienceworld} world modelling tasks, on Step 2 on qwen.}
\label{tab:scienceworld37}
\begin{tabular}{lcccc}
\toprule
\textbf{Task Type} & \textbf{EM} & \textbf{F1}
  & \textbf{BLEU} & \textbf{Entity}\\
\midrule
Boil & 0.0\% & 37.8\% & 3.2\% & 37.0\% \\
Chemistry & 14.8\% & 43.8\% & 15.5\% & 50.0\% \\
Conductivity & 21.7\% & 59.3\% & 26.6\% & 64.9\% \\
Find & 15.4\% & 56.5\% & 23.7\% & 70.5\% \\
Freeze & 12.5\% & 47.2\% & 24.7\% & 50.0\% \\
Genetics & 50.0\% & 69.0\% & 54.5\% & 69.7\% \\
Grow & 29.2\% & 57.6\% & 34.0\% & 64.7\% \\
Incline & 57.3\% & 88.3\% & 80.2\% & 88.0\% \\
Melt & 6.3\% & 30.0\% & 12.2\% & 32.8\% \\
Power & 14.3\% & 44.2\% & 16.5\% & 66.7\% \\

\bottomrule
\end{tabular}
\end{table}

\begin{table}[h]
\centering
    \setlength\tabcolsep{15pt}
    \setlength\extrarowheight{0pt}
\caption{The effect of deferred decoding on the ScienceWorld dataset \citep{wang-etal-2022-scienceworld} world modelling tasks, on Step 3 on qwen.}
\label{tab:scienceworld38}
\begin{tabular}{lcccc}
\toprule
\textbf{Task Type} & \textbf{EM} & \textbf{F1}
  & \textbf{BLEU} & \textbf{Entity}\\
\midrule
Boil & 0.0\% & 48.7\% & 17.6\% & 63.0\% \\
Chemistry & 11.1\% & 45.6\% & 16.4\% & 61.4\% \\
Conductivity & 26.1\% & 65.1\% & 30.0\% & 79.6\% \\
Find & 15.4\% & 46.7\% & 16.8\% & 59.0\% \\
Freeze & 0.0\% & 37.6\% & 8.3\% & 64.6\% \\
Genetics & 27.5\% & 59.7\% & 29.4\% & 61.1\% \\
Grow & 18.5\% & 50.3\% & 22.3\% & 57.1\% \\
Incline & 54.0\% & 89.3\% & 80.5\% & 90.1\% \\
Melt & 9.5\% & 34.8\% & 16.4\% & 42.8\% \\
Power & 14.3\% & 42.0\% & 18.1\% & 42.4\% \\

\bottomrule
\end{tabular}
\end{table}

\begin{table}[h]
\centering
    \setlength\tabcolsep{15pt}
    \setlength\extrarowheight{0pt}
\caption{The effect of deferred decoding on the ScienceWorld dataset \citep{wang-etal-2022-scienceworld} world modelling tasks, on Step 4 on qwen.}
\label{tab:scienceworld39}
\begin{tabular}{lcccc}
\toprule
\textbf{Task Type} & \textbf{EM} & \textbf{F1}
  & \textbf{BLEU} & \textbf{Entity}\\
\midrule
Boil & 0.0\% & 32.6\% & 0.0\% & 33.3\% \\
Chemistry & 14.8\% & 45.0\% & 19.3\% & 46.5\% \\
Conductivity & 30.4\% & 64.0\% & 31.0\% & 74.7\% \\
Find & 38.5\% & 61.1\% & 40.0\% & 71.8\% \\
Freeze & 12.5\% & 45.6\% & 20.8\% & 57.3\% \\
Genetics & 48.3\% & 67.0\% & 52.0\% & 69.0\% \\
Grow & 35.4\% & 62.5\% & 39.1\% & 70.7\% \\
Incline & 59.3\% & 87.5\% & 80.0\% & 88.1\% \\
Melt & 7.9\% & 30.7\% & 12.9\% & 36.0\% \\
Power & 0.0\% & 38.9\% & 7.2\% & 63.1\% \\

\bottomrule
\end{tabular}
\end{table}

\begin{table}[h]
\centering
    \setlength\tabcolsep{15pt}
    \setlength\extrarowheight{0pt}
\caption{The effect of deferred decoding on the ScienceWorld dataset \citep{wang-etal-2022-scienceworld} world modelling tasks, on Step 5 on qwen.}
\label{tab:scienceworld40}
\begin{tabular}{lcccc}
\toprule
\textbf{Task Type} & \textbf{EM} & \textbf{F1}
  & \textbf{BLEU} & \textbf{Entity}\\
\midrule
Boil & 11.1\% & 50.3\% & 26.5\% & 41.5\% \\
Chemistry & 11.1\% & 44.9\% & 13.9\% & 50.2\% \\
Conductivity & 26.1\% & 63.8\% & 30.1\% & 79.5\% \\
Find & 30.8\% & 60.0\% & 39.6\% & 79.4\% \\
Freeze & 0.0\% & 36.6\% & 9.3\% & 47.7\% \\
Genetics & 38.3\% & 62.4\% & 41.2\% & 65.9\% \\
Grow & 29.2\% & 60.0\% & 33.5\% & 66.3\% \\
Incline & 54.0\% & 89.0\% & 80.0\% & 89.1\% \\
Melt & 7.9\% & 40.3\% & 18.4\% & 43.2\% \\
Power & 14.3\% & 36.3\% & 22.1\% & 41.5\% \\

\bottomrule
\end{tabular}
\end{table}

Across the tables, we conclude that the deferred decoding is useful, and it tends to be more useful when the rollouts are accumulated.
\begin{figure*}[t]
\centering
\includegraphics[width=\textwidth]{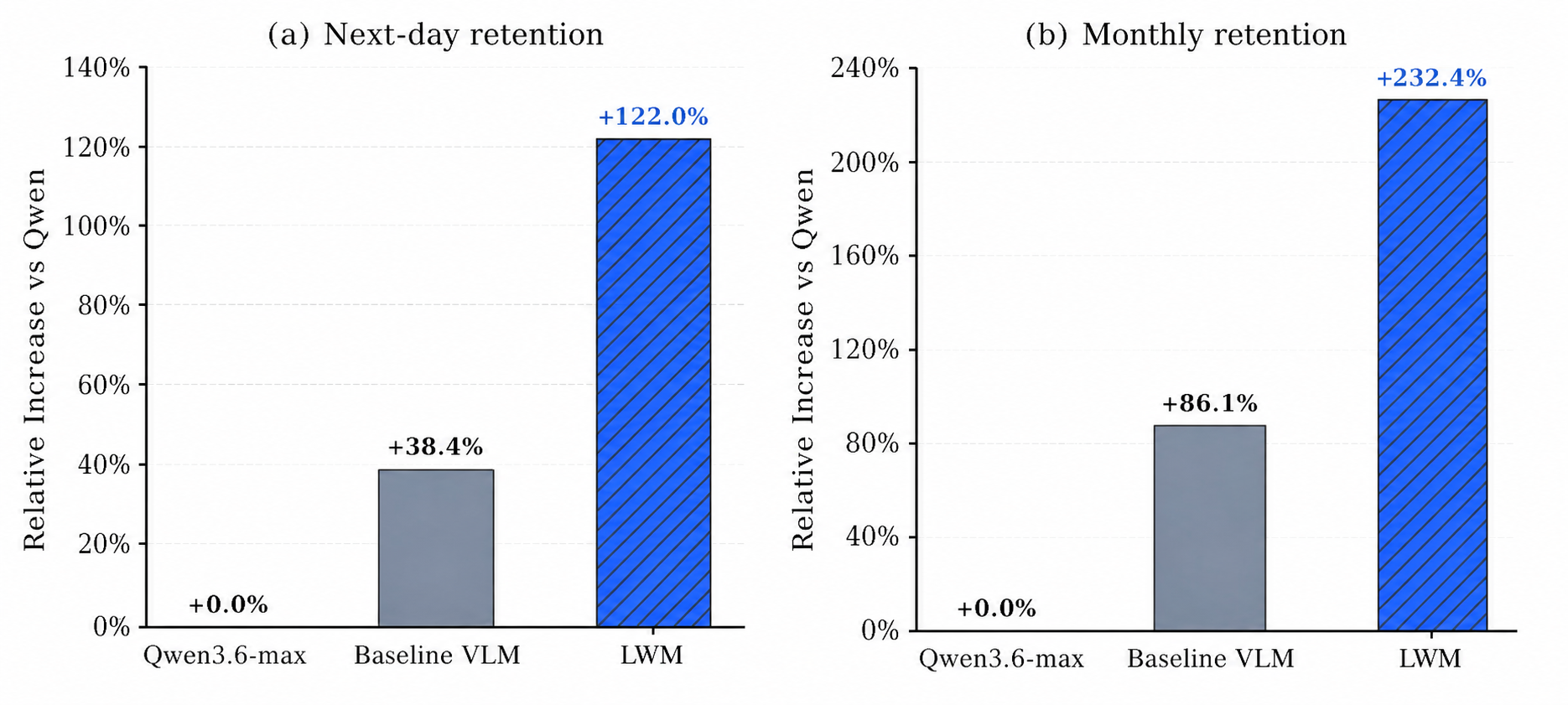}
\caption{Relative increase over Qwen3.7-max on automatic online performance, compared against baselines. Note that the results are obtained via online estimation, with the tasks of danmaku generation. LWM denotes LoopWM.}
\label{fig: compare1}
\end{figure*}

\begin{figure*}[t]
\centering
\includegraphics[width=\textwidth]{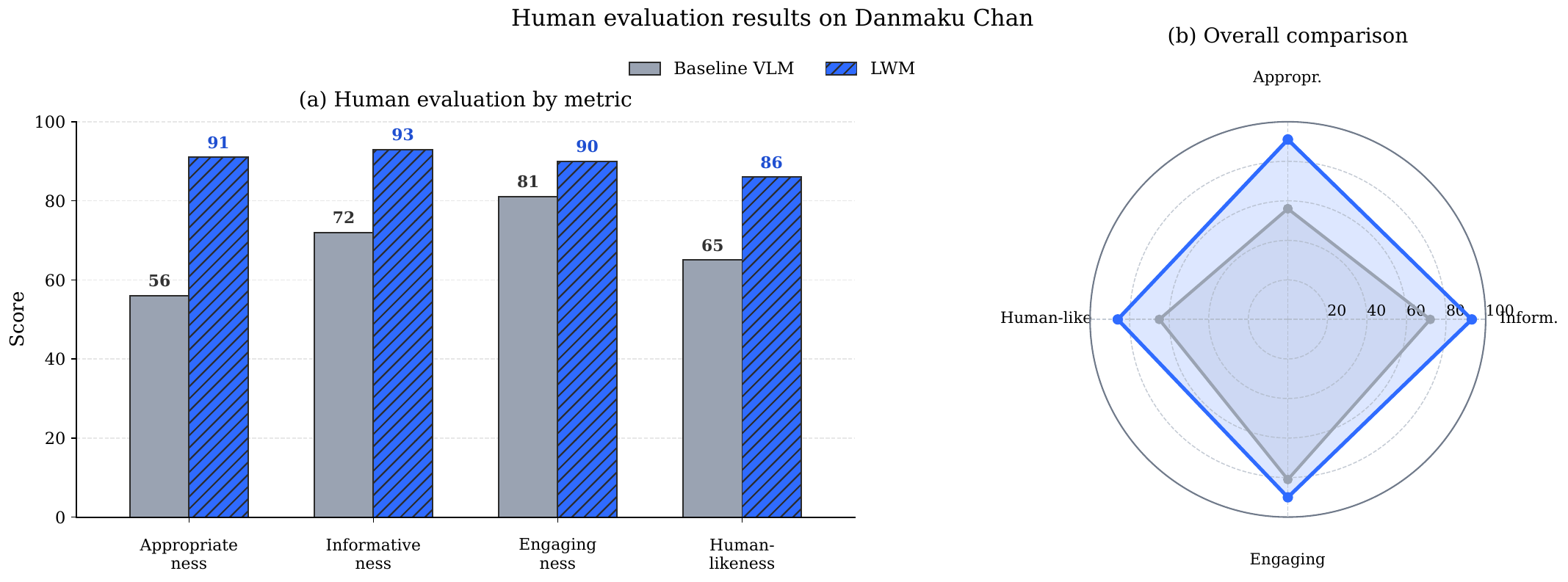}
\caption{Human evaluation performance with our model, compared against baselines. Note that the results are obtained via online estimation, with the tasks of danmaku generation. LWM denotes LoopWM.}
\label{fig: method2}
\end{figure*}

\section{Conclusions}
We have presented Looped World Models, the first application of looped transformer architectures to world modelling. Our approach addresses a central tension in current world models: generating faithful long-horizon simulations demands deep computation, yet deeper models incur prohibitive deployment costs and are susceptible to compounding rollout errors. By iteratively refining latent environment states through a parameter-shared transformer block with stabilised residual dynamics, LoopWM structurally mirrors the recurrence inherent in physical systems while maintaining a compact parameter footprint. Empirically, LoopWM achieve up to 100× parameter efficiency over conventional approaches without sacrificing prediction quality. Theoretically, we show that spectral-norm constraints on state transitions yield provably stable rollouts, providing formal guarantees that are absent in standard autoregressive world models. Furthermore, our adaptive computation mechanism automatically scales the effective depth of the model to match the complexity of each prediction step, allocating more refinement iterations to dynamically challenging transitions and fewer to predictable ones. Beyond the specific results reported here, we believe this work identifies iterative latent depth as a new scaling axis for world simulation, one that is orthogonal to the conventional axes of model size and data volume. We hope that this perspective opens new directions for building world models that are simultaneously more capable, more efficient, and more stable over extended horizons.

\section{Broader Impacts}
While the present paper already provides strong evidence for the effectiveness of LoopWM, the current manuscript is intentionally selective in disclosure scope. In this version, our goal is to establish the core architectural thesis that looped latent refinement, deferred decoding, and stabilized dynamics together define a viable and promising design space for world modelling, rather than to exhaustively present every supporting result we have already obtained.
\par
First, the current paper already demonstrates the value of iterative latent computation through deferred decoding, which gives concrete evidence that preserving and refining latent computation across rollout steps is beneficial. We view this as a direct and meaningful manifestation of the looped design. At the same time, it represents only one visible entry point into a broader body of evidence supporting the effectiveness of looping, and a more explicit decomposition of these gains can be disclosed in the future.
\par
Second, although the present manuscript emphasizes the principal task domains reported here, our empirical validation is not confined to these settings. We have also verified in continuous visual environments that optimization is feasible and that the training loss is consistently reducible, which supports the practicality of the proposed architecture beyond the environments highlighted in this paper. The main limitation at this stage is therefore not a lack of empirical support, but that the manuscript does not yet fully expose the breadth of validation already completed.
\par
Third, LoopWM is best understood as a distinct point in the broader world model landscape. The current paper makes clear that its emphasis differs from major existing families, including RSSM style latent dynamics models, autoregressive video token world models, and diffusion based world models. A more explicit positioning analysis would further sharpen this distinction and make the contribution even easier to interpret. We therefore see clear value in more directly situating LoopWM among these families and clarifying the regimes in which iterative latent depth is the most natural scaling axis.
\par
Finally, our current step 1 to step 5 experiments already indicate that iterative latent depth behaves as a meaningful scaling dimension, and we consider this one of the central implications of the work. The remaining limitation is not whether such a scaling trend exists, but that the present paper stops short of providing a more complete scaling law characterization across broader task and compute ranges. Similarly, from an optimization perspective, our experience suggests that training can benefit from curriculum like engineering strategies that progressively unlock the architecture’s capability. We regard this not as a weakness of the method, but as part of the practical recipe for making a new architectural regime reliably trainable at scale.
\par
Overall, the main limitation of the current paper is one of presentation scope rather than conceptual or empirical foundation. The paper establishes the core case for LoopWM, while broader cross family positioning, richer scaling analysis, and more extensive optimization disclosure can further strengthen the story in the future.

\clearpage
\bibliographystyle{iclr2025_conference}  
\bibliography{iclr2025_conference}  
\end{document}